%% file: _main.tex
\documentclass[10pt,twocolumn,letterpaper]{article}

\usepackage{cvpr}              % To produce the CAMERA-READY version
\usepackage{graphicx}
\usepackage{grffile}
\usepackage[utf8]{inputenc} % allow utf-8 input
\usepackage[T1]{fontenc}    % use 8-bit T1 fonts
\usepackage{url}            % simple URL typesetting
\usepackage{booktabs}       % professional-quality tables
\usepackage{amsfonts}       % blackboard math symbols
\usepackage{nicefrac}       % compact symbols for 1/2, etc.
\usepackage{microtype}      % microtypography
\usepackage{xcolor}         % colors
\usepackage{adjustbox}

\usepackage{bbding}
\usepackage{multirow}
\usepackage{booktabs}
\usepackage{makecell}
\usepackage{bbm}
\usepackage{courier}
\usepackage{caption}
\usepackage{float}
\usepackage{mathtools}

\usepackage{colortbl}
\usepackage[pagebackref,breaklinks,colorlinks]{hyperref}

\definecolor{mygray}{gray}{.9}

\usepackage[capitalize]{cleveref}
\crefname{section}{Sec.}{Secs.}
\Crefname{section}{Section}{Sections}
\Crefname{table}{Table}{Tables}  
\crefname{table}{Tab.}{Tabs.}

\def\cvprPaperID{6347} % *** Enter the CVPR Paper ID here
\def\confName{CVPR}
\def\confYear{2023}

\newcommand{\modelname}{{Lite DETR }}
\newcommand{\prefix}{{Lite}}

\begin{document}

\title{\modelname: An Interleaved Multi-Scale Encoder for Efficient DETR}

\author{\textbf{Feng Li$^{1,2}$\thanks{This work was done when Feng Li was an intern at IDEA. }, ~Ailing Zeng$^{2}$, ~Shilong Liu$^{2,3}$, ~Hao Zhang$^{1,2}$, ~Hongyang Li$^{2,4}$} \\ \textbf{~Lei Zhang$^{2}$\thanks{Corresponding author.}\hspace{1.mm}, ~Lionel M. Ni$^{1,5}$} \\
$^1$The Hong Kong University of Science and Technology. \\
$^2$International Digital Economy Academy (IDEA). \\
$^3$Dept. of CST., BNRist Center, Institute for AI, Tsinghua University. \\
$^4$South China University of Science and Technology.\\
$^5$The Hong Kong University of Science and Technology (Guangzhou).\\
\texttt{\{fliay,hzhangcx\}@connect.ust.hk} \\
\texttt{\{liusl20\}@mails.tsinghua.edu.cn} \\
\texttt{\{eeli.hongyang\}@mail.scut.edu} \\
\texttt{\{leizhang\}@idea.edu.cn} \\
\texttt{\{ni\}@ust.hk} \\
}
\maketitle

\begin{abstract}
% Multi-scale features are of significant importance for object detection. However, the excessive token number in multi-scale features, especially for low-level features, is computationally inefficient for recent Transformer-based detection models.
% In this paper, we present Light DETR, a lightweight end-to-end object detection framework that effectively compresses the feature tokens by compressing informative tokens from low-level feature maps. More specifically, we propose an efficient encoder block to compress the abundant and informative low-level features into high-level features to construct compressed features in the encoder. At the end of each block, we use a feature expansion module to extricate the low-level features from the reduced features. In addition, to better compress local details from the low-level features,
% we introduce key-aware deformable attention. As a result, we significantly reduce the detection head GFLOPs by 60\% while keeping 99\% of the original performance. Meanwhile, comprehensive experiments validate the proposed simple and light encoder can generalize well across many DETR-based models.
% We will release the code after the blind review.

Recent DEtection TRansformer-based (DETR) models have obtained remarkable performance. Its success cannot be achieved without the re-introduction of multi-scale feature fusion in the encoder.
However, the excessively increased tokens in multi-scale features, especially for about 75\% of low-level features, are quite computationally inefficient, which hinders real applications of DETR models.
In this paper, we present Lite DETR, a simple yet efficient end-to-end object detection framework that can effectively reduce the GFLOPs of the detection head by 60\% while keeping 99\% of the original performance.
Specifically, we design an efficient encoder block to update high-level features (corresponding to small-resolution feature maps) and low-level features (corresponding to large-resolution feature maps) in an interleaved way.
In addition, to better fuse cross-scale features, we develop a key-aware deformable attention to predict more reliable attention weights. 
Comprehensive experiments validate the effectiveness and efficiency of the proposed Lite DETR, and the efficient encoder strategy can generalize well across existing DETR-based models.
The code will be available in \url{https://github.com/IDEA-Research/Lite-DETR}.

\end{abstract}

\input{resources/sec/intro}

\input{resources/sec/related}
\input{resources/sec/method}
\input{resources/sec/exp}
\input{resources/sec/conclusion}

{\small
\bibliographystyle {ieee_fullname}
\bibliography{egbib}
}

\end{document}

\title{---Supplementary Materials---\\ \modelname: An Interleaved Multi-Scale Encoder for Efficient DETR}

% \and
\maketitle

% {\small
% \bibliographystyle {ieee_fullname}
% \bibliography{egbib}
% }
\appendix
%\section{Visualization Analysis}
% In this section,
In this supplementary material, we share more qualitative analyses compared with other related works and verify our design choices that are not presented in our main paper, including:
\begin{itemize}
    \item Comparisons between sparse DETR and Lite deformable DETR in Sec.~\ref{sec:supp1}.
    \item Comparisons between DINO-3scale and Lite DINO in Sec.~\ref{sec:supp2}. 
    \item Comparisons between deformable attention and KDA attention in Sec.~\ref{sec:supp3}. 
    \item More visualization of sampled locations between deformable attention and KDA attention in Sec.~\ref{sec:supp4}.
    \item Our failure cases in Sec.~\ref{sec:supp5}.
    % \item Limitation of our work in Sec.~\ref{sec:limitation}.
    % \item Broader impact in Sec.~\ref{sec:impact}.
    % \item Code submission and reproducibility in Sec.~\ref{sec:code}.
\end{itemize}
All the models are based on Deformable DETR and DINO, which are denoted as Lite-Deformable DETR and Lite DINO. Except for models in Sec. \ref{sec:sparse} that are based on Deformable DETR, other models are based on DINO.
Please note that all boxes shown in these figures are selected from predicted boxes of the corresponding models with classification scores larger than $0.3$.

\section{Comparison between Sparse DETR and Lite Deformable DETR}\label{sec:sparse}
\label{sec:supp1}
\begin{figure*}[hb]
    \centering
    \includegraphics[width=1\textwidth]{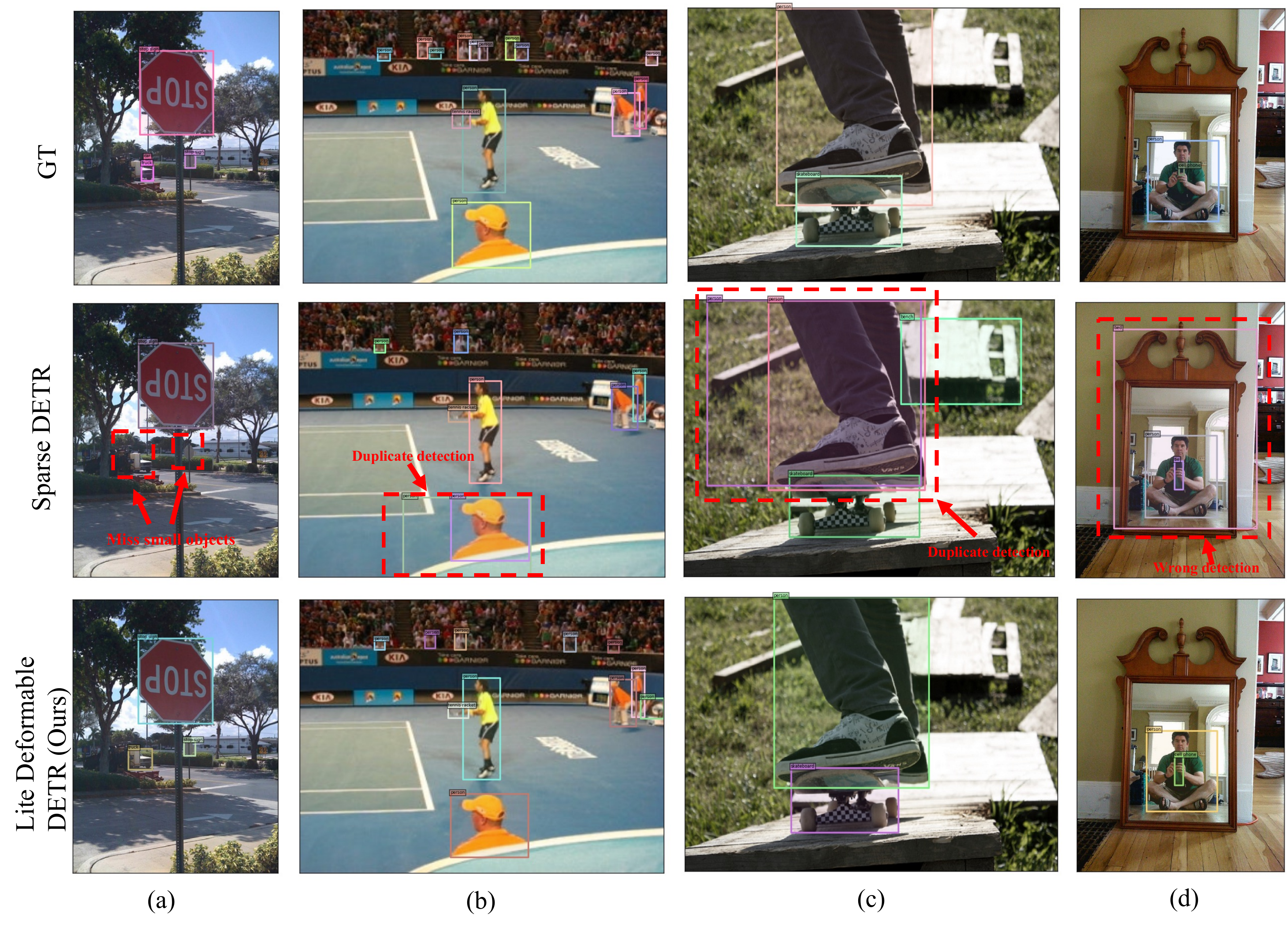}
    % \vspace{-5.3cm}
    \caption{Visualization of detection results in Sparse DETR and our Lite-Deformable DETR. (a) shows that Sparse DETR may miss small objects in some cases. (b), (c) and (d) demonstrate that Sparse DETR is inferior to our model in medium and large objects, where it tends to predict duplicate and wrong boxes.}
    \label{fig:supp_compare_sparsedetr}
\end{figure*}

\input{resources/tables/sparselargecompare}

To further analyze why our Lite-Deformable DETR outperforms Sparse DETR~\cite{roh2021sparse}, we conduct a visualization of these two models in Fig.~\ref{fig:supp_compare_sparsedetr}.  In Fig.~\ref{fig:supp_compare_sparsedetr}(a), we show that Sparse DETR may miss small objects in some cases. More importantly, in Fig.~\ref{fig:supp_compare_sparsedetr}(b), (c), and (d), we demonstrate that Sparse DETR is inferior to our model in medium and large objects in that it tends to predict duplicate and wrong boxes. This phenomenon is also consistent with the results in Table~\ref{tab:sparsecompare}, i.e., our model outperforms Sparse DETR by 1.6 AP in $AP_L$ under comparable GFLOPs. As Sparse DETR only selects some tokens from multi-scale features, it breaks the structured feature organization, especially for high-level features with rich semantics. Therefore, it impacts large object detection and is difficult to plug in existing detection models as a general strategy.

\section{Comparison between DINO-3scale and Lite DINO}
\label{sec:supp2}
\begin{figure*}[h]
    \centering
    \includegraphics[width=1\textwidth]{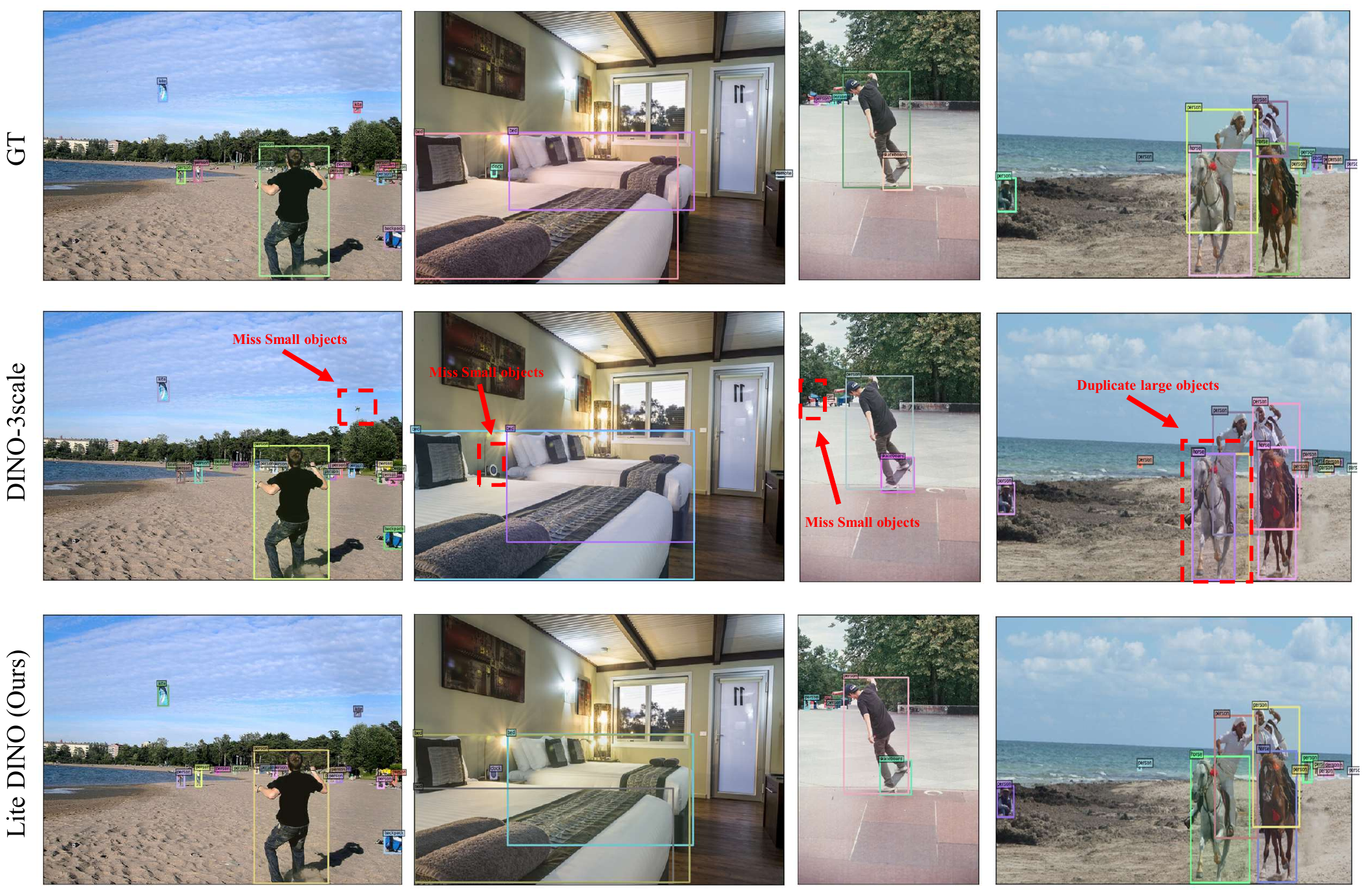}
    \caption{Visualization of detection results in DINO-3scale and our proposed Lite DINO. Directly dropping the high-resolution map (DINO-3scale) will make small object detection difficult, while Lite DINO can maintain comparable performance as the original DINO.}
    \label{fig:supp_drop_4scale}
\end{figure*}

As we claimed in the main paper, the high-resolution (low-level) map is redundant but important and should be preserved properly. Simply dropping these features will harm the performance of small object detection. In Fig.~\ref{fig:supp_drop_4scale}, we present the visualization comparisons of directly dropping the high-resolution map (DINO-3scale) and our proposed Lite DINO. Detecting small objects would be more difficult for DINO-3scale. Though Lite DINO slightly increases the GFLOPs compared to DINO-3scale, it maintains comparable performance as the original DINO. 

\section{Comparison between Deformable and KDA Attention}
\label{sec:supp3}
\begin{figure*}[h]
    \centering
    \includegraphics[width=1\textwidth]{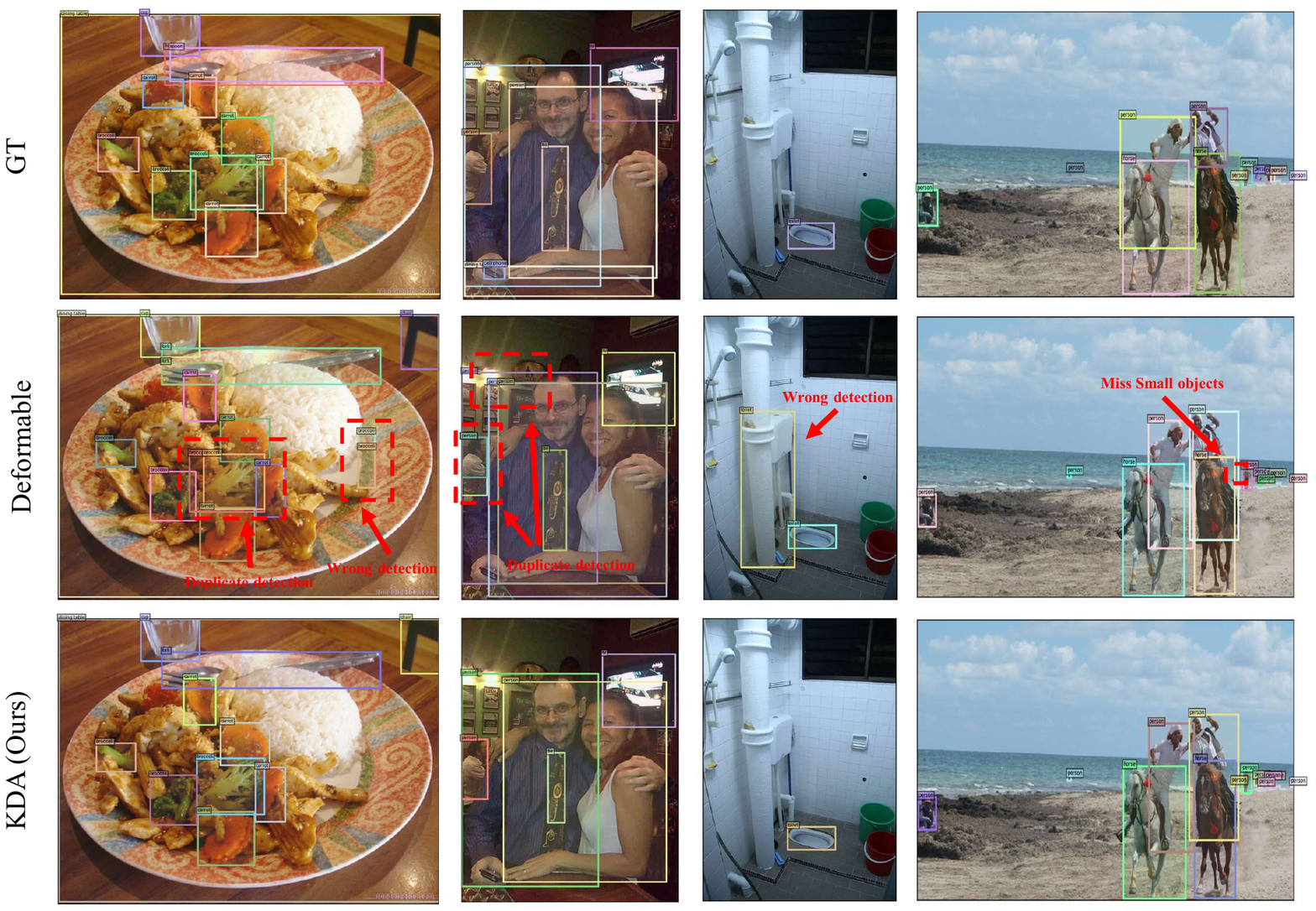}
    \caption{Visualization of detection results in our Lite DINO by using deformable attention and the proposed KDA attention. KDA attention shows its superiority in improving small object detection and reducing duplicate detection.}
    \label{fig:supp_use_kda}
\end{figure*}

To enhance the lagged low-level feature update in the Lite DETR framework, we visualize how our KDA attention outperforms the original deformable attention in Fig.~\ref{fig:supp_use_kda}. KDA attention shows its superiority in improving small object detection and reducing duplicate detection, indicating it can mitigate the effects of asynchronous features exposed in the Lite DETR framework.

\section{More Visualization of Sampled Locations between Deformable and KDA Attention}
\label{sec:supp4}
\begin{figure*}[h]
    \centering
    \includegraphics[width=1\textwidth]{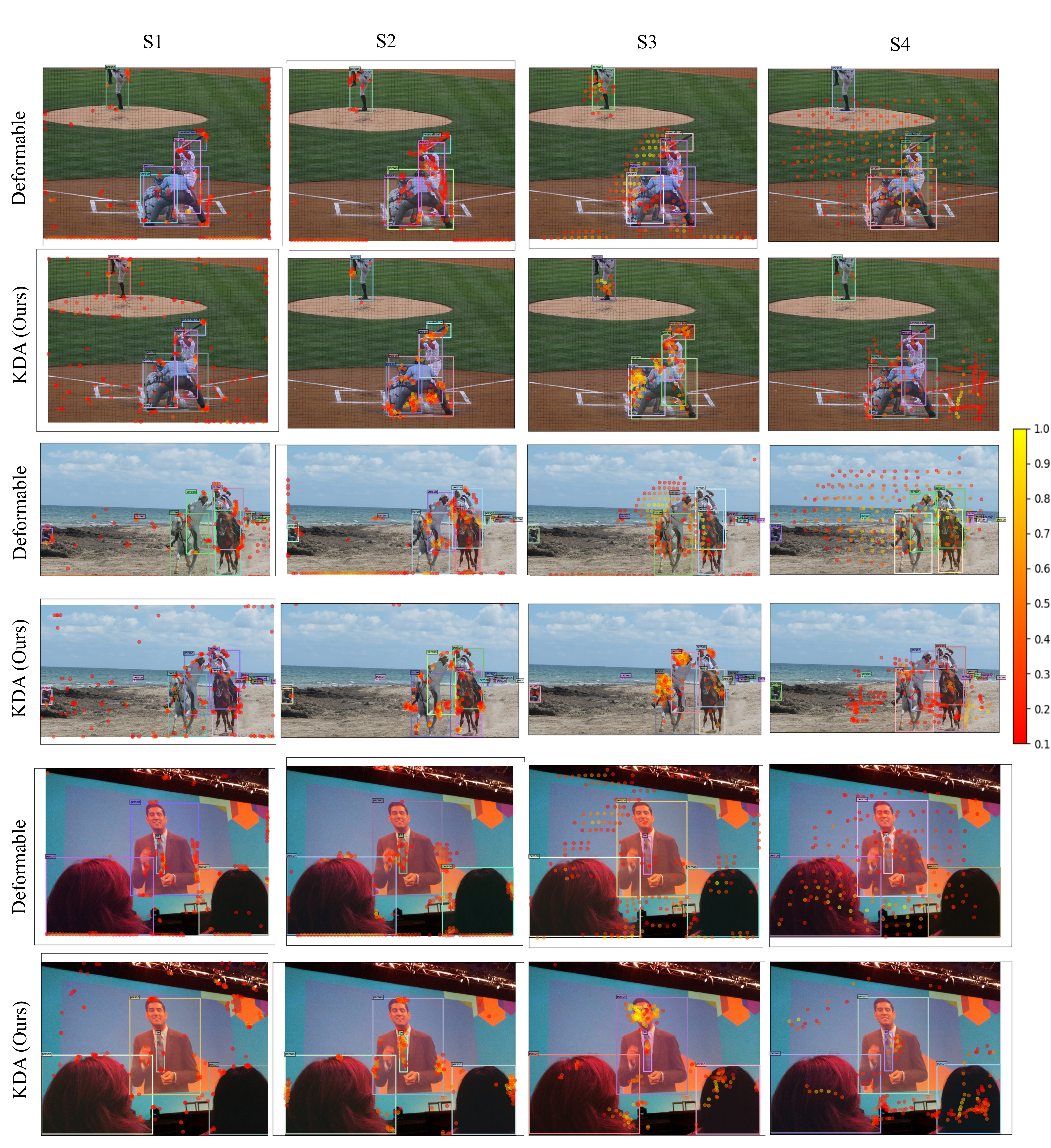}
    \caption{Visualization comparison of deformable and KDA attention in our Lite DINO from all feature scales. In high-level feature maps (S1, S2), KDA and deformable attention show similar attention regions. However, kDA can sample more meaningful locations on low-level maps (S3, S4), which indicates KDA can better extract local features from low-level maps and improve average precision in small object detection.}
    \label{fig:supp_attn1}
\end{figure*}

\begin{figure*}[h]
    \centering
    \includegraphics[width=1\textwidth]{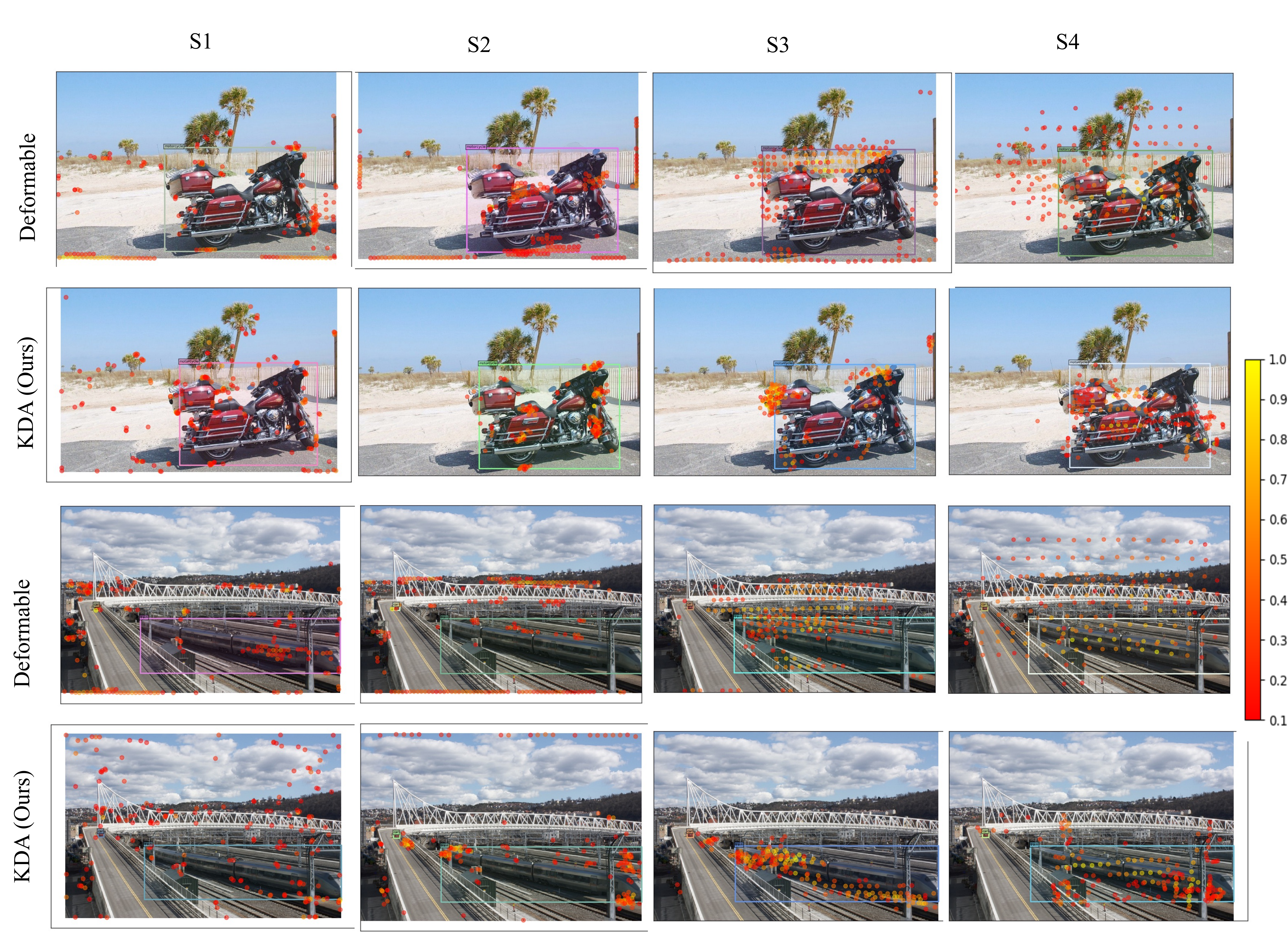}
    \caption{Visualization comparison of deformable and KDA attention in our Lite DINO from all feature scales. }
    \label{fig:supp_att2}
\end{figure*}

In Fig. 5 in our paper, we provide a few visualization maps of deformable and KDA attention under Lite DINO. To better show their differences, we further provide more attention to visualization results in our interleaved encoder. Following Fig. 5 in our paper, we show the top 200 sampling locations on all scales (S1, S2, S3, and S4) for all query tokens in Fig. \ref{fig:supp_attn1} and \ref{fig:supp_att2}. In high-level feature maps (S1, S2), KDA and deformable attention focus on similar regions. However, kDA can sample more meaningful locations on low-level maps with high resolution (S3, S4), which indicates KDA can better extract local features from low-level maps and improve average precision in small object detection.

\section{Failure Cases in Lite DETR}
\label{sec:supp5}
\begin{figure*}[h]
    \centering
    \includegraphics[width=1\textwidth]{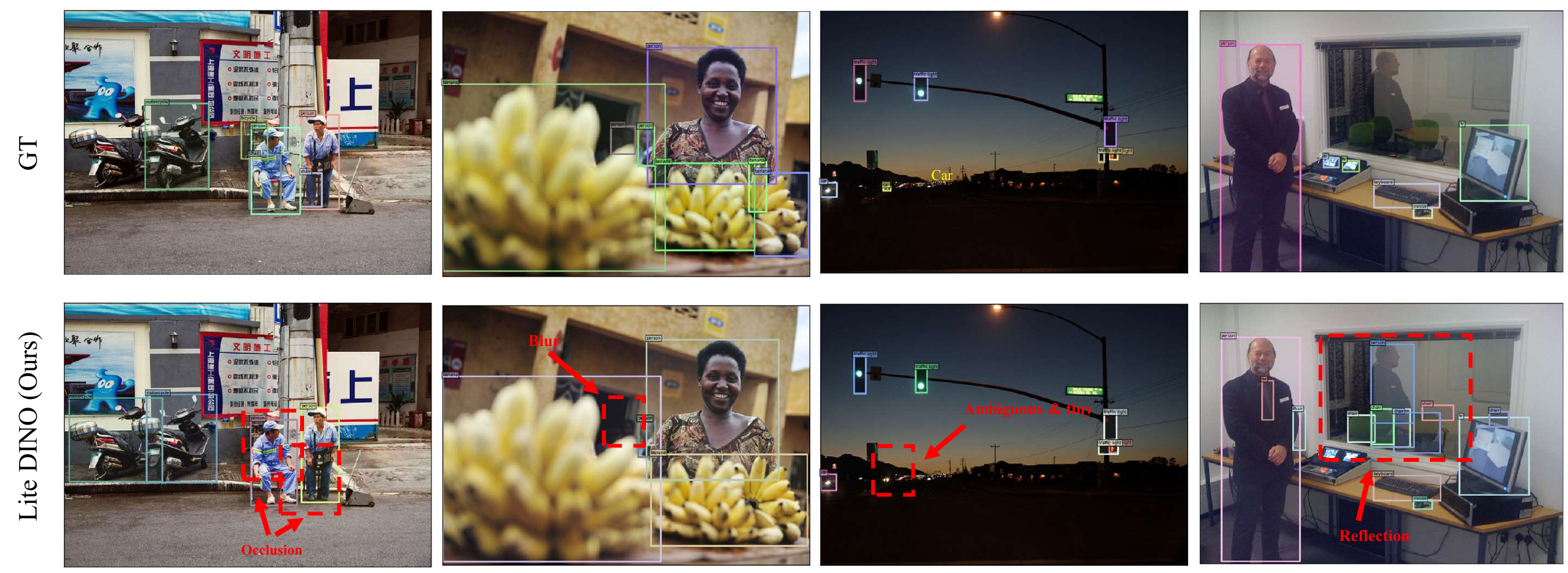}
    \caption{Visualization of failure cases in our Lite DINO. }
    \label{fig:supp_failure}
\end{figure*}

In Fig.~\ref{fig:supp_failure}, we analyze the cases when our method fails, including occlusion, blur, ambiguity, and reflection. These cases are quite difficult to detect, even for a human, and we will leave it for future work to address these cases.

% \clearpage

{\small
\bibliographystyle {ieee_fullname}
\bibliography{egbib}
}

%% file: resources/sec/intro.tex
\section{Introduction}
Object detection aims to detect objects of interest in images by localizing their bounding boxes and predicting the corresponding classification scores. In the past decade, remarkable progress has been made by many classical detection models \cite{ren2015faster, redmon2018yolov3} based on convolutional networks. Recently, DEtection TRansformer \cite{carion2020end} (DETR) introduces Transformers into object detection, and DETR-like models have achieved promising performance on many fundamental vision tasks, such as object detection \cite{zhu2020deformable,zhang2022dino,li2022dn}, instance segmentation \cite{cheng2021maskformer, cheng2021mask2former,li2022mask}, and pose estimation~\cite{stoffl2021end,shi2022end}.

\input{resources/images/introimage}
Conceptually, DETR \cite{carion2020end} is composed of three parts: a backbone, a Transformer encoder, and a Transformer decoder. Many research works have been improving the backbone and decoder parts. For example, the backbone in DETR is normally inherited and can largely benefit from a pre-trained classification model \cite{liu2021swin, he2015deep}. The decoder part in DETR is the major research focus, with many research works trying to introduce proper structure to DETR query and improve its training efficiency~\cite{zhu2020deformable, meng2021conditional, li2022dn, liu2022dab, zhang2022dino,jia2022detrs}. By contrast, much less work has been done to improve the encoder part. 
% to produce features of different scales. The backbone features usually establish a multi-scale feature pyramid.
%
The encoder in vanilla DETR includes six Transformer encoder layers, stacked on top of a backbone to improve its feature representation. Compared with classical detection models, it lacks multi-scale features, which are of vital importance for object detection, especially for detecting small objects~\cite{lin2017feature, tan2020efficientdet, liu2018path, ghiasi2019fpn, qiao2021detectors}. 
% Actually, encoder in DETR is similar to the FPN \cite{lin2017feature} in CNN-based models, which will fuse and enhance the multi-scale features for the decoder. 
%
% In fact, multi-scale features are of vital importance for object detection, especially small objects, which has been widely discussed in many previous CNN-based detectors \cite{lin2017feature, tan2020efficientdet, liu2018path, ghiasi2019fpn, qiao2021detectors}.
%
Simply applying Transformer encoder layers on multi-scale features is not practical due to the prohibitive computational cost that is quadratic to the number of feature tokens. 
For example, DETR uses the C5 feature map, which is 1/32 of the input image resolution, to apply the Transformer encoder. If a C3 feature (1/8 scale) is included in the multi-scale features, the number of tokens from this scale alone will be 16 times of the tokens from the C5 feature map. The computational cost of self-attention in Transformer will be 256 times high. 
%
%
% To achieve high detection performance, many previous works \cite{meng2021conditional,li2022dn, liu2022dab, zhang2022dino,yao2021efficient} have attempted to improve the \emph{decoder} from various aspects, such as decoupling content and positional information \cite{meng2021conditional}, providing better formulations of positional queries \cite{liu2022dab, zhu2020deformable}, utilizing dense priors from the encoder \cite{yao2021efficient, zhu2020deformable, zhang2022dino} for better decoder query initialization, and improving the one-to-one label-assignment for fast convergence \cite{li2022dn, zhang2022dino, jia2022detrs, chen2022group}.
% %
% These efforts continued to push the performance of object detection up until DINO \cite{zhang2022dino} reached first place on COCO2017 object detection, outperforming classical detectors for the first time.
%These works have achieved remarkable performance, for example, DINO \cite{zhang2022dino} achieved first place on COCO2017 object detection, outperforming classical detectors for the first time.

%As self-attention in Transformers has quadratic complexity with the token number, Vanilla DETR \cite{carion2020end} is not capable of using multi-scale features due to the prohibitive computational cost in the encoder. 
% As the performance of these methods rises, we find that the high computational overhead issue becomes non-negligible.
%
To address this problem, Deformable DETR~\cite{zhu2020deformable} develops a deformable attention algorithm to reduce the self-attention complexity from quadratic to linear by comparing each query token with only a fixed number of sampling points. 
%Similarly, Dynamic Head~\cite{dai2021dynamic} also proposes to use deformable convolution to reduce the complexity of spatial-aware attention. 
Based on this efficient self-attention computation, Deformable DETR introduces multi-scale features to DETR, and the deformable encoder has been widely adopted in subsequent DETR-like models~\cite{li2022dn, liu2022dab, zhang2022dino,jia2022detrs}.

However, 
%the computational cost is still large for many applications. The main reason for this 
due to a large number of query tokens introduced from multi-scale features, the deformable encoder still suffers from a high computational cost.
To reveal this problem, we conduct some analytic experiments as shown in Table~\ref{tab:dinoflops} and ~\ref{tab:feature scale} using a DETR-based model DINO~\cite{zhang2022dino} to analyze the performance bottleneck of multi-scale features.  
Some interesting results can be observed. First, 
% multi-scale features contain many redundant tokens, especially 
the low-level (high-resolution map) features account for more than $75\%$ of all tokens. Second, direct dropping some low-level features (DINO-3scale) mainly affects the detection performance for small objects (AP\_S) by a 10\% drop but has little impact on large objects (AP\_L).

Inspired by the above observations, we are keen to address a question: \emph{can we use fewer feature scales but maintain important local details?}
Taking advantage of the structured multi-scale features, we present an efficient DETR framework, named \modelname. 
%It processes more compact features of high-level.
%
Specifically, we design a simple yet effective encoder block including several deformable self-attention layers, which can be plug-and-play in any multi-scale DETR-base models to reduce $62\%\sim78\%$ encoder GFLOPs and maintain competitive performance. 
The encoder block splits the multi-scale features into high-level features (e.g., C6, C5, C4) and low-level features (e.g., C3). 
High-level and low-level features will be updated in an interleaved way to improve the multi-scale feature pyramid.  That is, in the first few layers, we let the high-level features query all feature maps and improve their representations, but keep low-level tokens intact. 
Such a strategy can effectively reduce the number of query tokens to $5\%\sim25\%$ of the original tokens and save a great amount of computational cost. At the end of the encoder block, we let low-level tokens query all feature maps to update their representations, thus maintaining multi-scale features. In this interleaved way, we update high-level and low-level features in different frequencies for efficient computation.

% At the end of the encoder block, we then \emph{extricate} the low-level tokens to query all feature maps to update their representations without any dense self-attention operations.
% %
% Such a simple hide-and-extricate strategy can effectively reduce the number of query tokens participating quadratic self-attention operations and save a great amount of computational costs. 

%Specifically, we design an efficient encoder block, which conducts token reduction to reduce the number of encoder queries. This encoder block splits the multi-scale features into high-level features and low-level features. The high-level features are from low-resolution feature maps (e.g. C6, C5, C4) whereas the low-level features are from high-resolution feature maps (e.g. C3). The encoder block includes several deformable self-attention layers. Each layer only lets high-level tokens query all feature maps and improve their representations, but keeps low-level tokens untouched. Such a strategy can effectively reduce the number of query tokens to $5\%-25\%$ of the original tokens and save a great amount of computational cost. At the end of the encoder block, we let low-level tokens to query all feature maps to update their representations. 

Moreover, to enhance the lagged low-level feature update, we propose a key-aware deformable attention (KDA) approach to replacing all attention layers. When performing deformable attention, for each query, it samples both keys and values from the same sampling locations in a feature map. Then, it can compute more reliable attention weights by comparing the query with the sampled keys. Such an approach can also be regarded as an extended deformable attention or a sparse version of dense attention. 
% KDA slightly increases the computation cost of deformable attention. However, the increased cost is negligible compared with the saved computation cost by token reduction. 
We have found KDA very effective in bringing the performance back with our proposed efficient encoder block.

% Specially,  
% split the feature pyramid into high-level and low-level 
% We utilize the high-level tokens to extract important local features from the excessive low-level tokens, which builds compact tokens with only $5\%-25\%$ of the original tokens. These compact tokens contain both rich semantics and local details. We further propose a compressed attention block to enhance the compact tokens, which is composed of a self-attention to strengthen itself and a cross-attention with low-level features to fuse more local features.
% %
% %
% At the end of each encoder block, we will utilize the original low-level feature to query the local details back from the compact tokens, which efficiently extricates the multi-scale feature map back. 
% %
% To effectively fuse features between different scales in our efficient encoder block, we propose a Key-aware Deformable Attention (KDA). It samples both key and value from the feature map so that a query can decide the attention weight of each value by comparing it with the keys, which produces better attention weights.

To summarize, our contributions are as follows. 
\begin{itemize}
\vspace{-.2cm}
    \item We propose an efficient encoder block to update high-level and low-level features in an interleaved way, which can significantly reduce the feature tokens for efficient detection. This encoder can be easily plugged into existing DETR-based models.
    \vspace{-.2cm}
    \item To enhance the lagged feature update, we introduce a key-aware deformable attention for more reliable attention weights prediction.
    \vspace{-.2cm}
    \item Comprehensive experiments show that \modelname can reduce the detection head GFLOPs by $60\%$ and maintain $99\%$ detection performance. Specifically, our \prefix-DINO-SwinT achieves $53.9$ AP with $159$ GFLOPs.
\end{itemize}

%% file: resources/images/introimage.tex
\begin{figure}
    \centering
    % \vspace{-0.2cm}
    \includegraphics[scale=0.48]{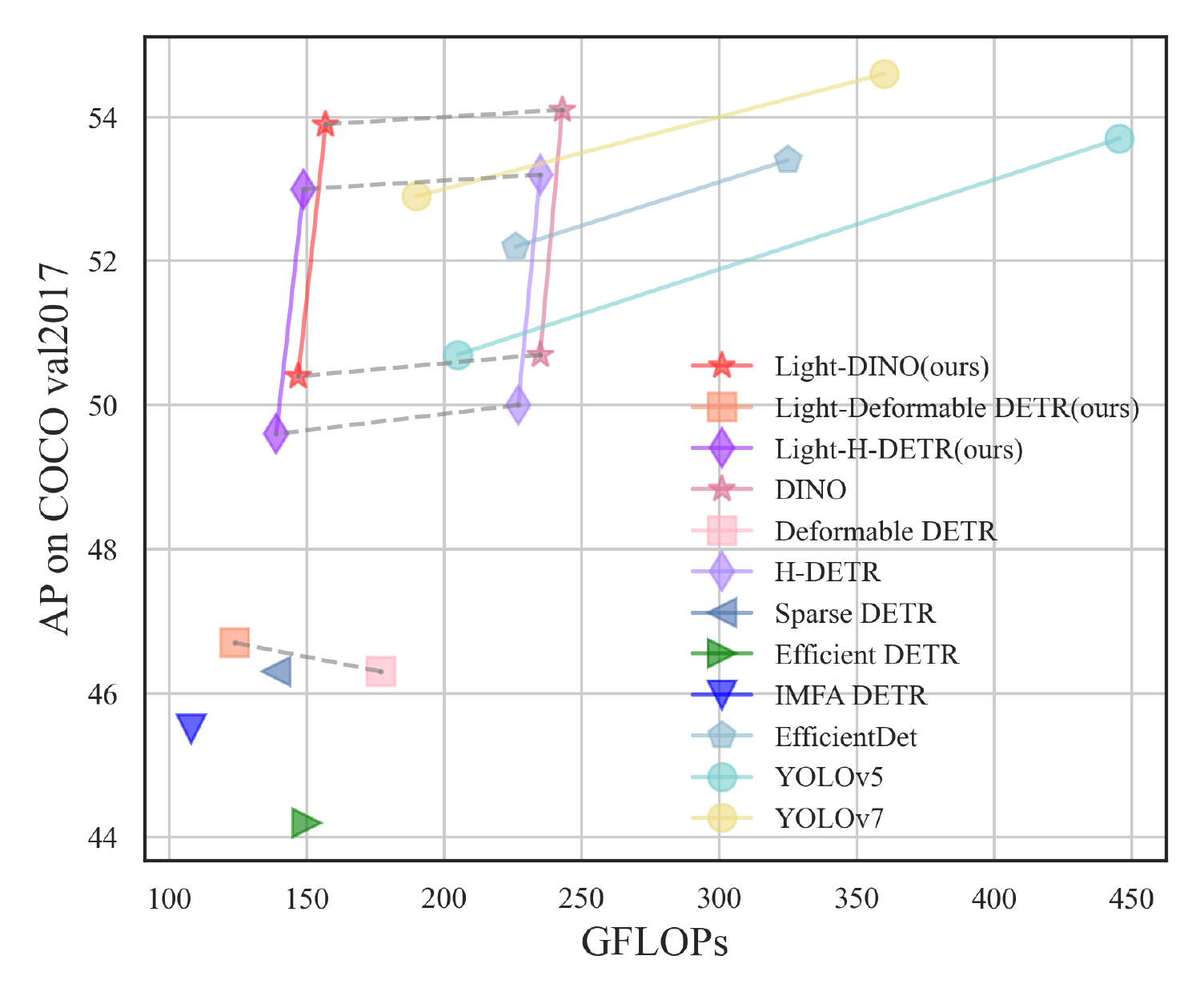}
    \vspace{-0.6cm}
    \caption{Average precision (Y axis) versus GFLOPs (X axis) for different detection models on COCO without extra training data. All models except EfficientDet \cite{tan2020efficientdet} and YOLO series \cite{wang2022yolov7,yolov5} use ResNet-50 and Swin-Tiny as backbones. Specifically, two markers on the same line use ResNet-50 and Swin-Tiny, respectively. Individual markers only use ResNet-50. Each dashed line connects algorithm variants before and after adding our algorithm. The size of the listed models vary from 32M to 82M.}
    \vspace{-0.5cm}
    \label{fig:converge}
\end{figure}

%% file: resources/sec/related.tex
\section{Related Work}
\input{resources/tables/flopsmem}
\noindent\textbf{Preliminary: }
DETR \cite{carion2020end} regards object detection as a direct set prediction problem and uses a set-based global loss to force unique predictions via bipartite matching. 
%
% Accordingly, different from previous classical detectors (e.g., Faster R-CNN~\cite{RenHG017}), it can be trained in an end-to-end manner without any manual post-processing.
% %
% In terms of the architecture, it is composed of a backbone, a Transformer encoder, and a Transformer decoder. The backbone is a pre-trained classification model that produces multi-scale features. 
% %
% Each feature scale is the output of an intermediate backbone stage.
%
Vanilla DETR \cite{carion2020end} only uses single-scale features from the last stage of the backbone ($\frac{1}{32}$ of the input image resolution), i.e., $X_{feat}\in R^{N\times D}$, where $D$ is the feature dimension and $N$ is the total number of flattened features. 
These features will then be processed by the encoder with dense self-attention layers for feature fusion and enhancement.
The use of an encoder is similar to FPN \cite{lin2017feature} in CNN-based models. 
%
%The decoder is a detection head, which will query the enhanced features via learnable decoder queries to detect objects.
%
These refined features will be queried by the decoder to detect objects by predicting classification scores and regressing bounding boxes.
In general, as DETR only uses high-level features that are of low resolution, 
% (1/32 of the input image resolution), 
these features lack rich local details that are critical for small object detection. 
%
% As a result, a large number of improvements over DETR have emerged.
%It is an end-to-end object detector based on encoder-decoder Transformers. It considers object detection as a set-based prediction task and uses learnable decoder queries to enhance the encoded features from the encoder. Generally, DETR is composed of a backbone, a Transformer encoder, and a Transformer decoder. The backbone is a pretrained classification model that produces multi-scale features. These features will then be processed by the encoder for feature enhancement. Therefore, the encoder is similar to the FPN \cite{lin2017feature} in CNN-based models. The decoder is a detection head, which will query the enhanced features to detect objects.
\noindent\textbf{Improving Decoder Design of DETR:}
Recently, DETR-based detectors have seen more rapid progress~\cite{meng2021conditional,li2022dn, liu2022dab, zhang2022dino,yao2021efficient} compared to classical detectors~\cite{ren2015faster, chen2019hybrid}.
As a result, DINO \cite{zhang2022dino} achieved first place in COCO 2017 object detection for the first time as a DETR-like model.
Most works focus on improving the Transformer decoder in DETR for better performance and faster convergence speed.
% Before that, many works attempted to improve the Transformer decoder to achieve good detection performance. 
%
Specifically, Meng \textit{et al.} \cite{meng2021conditional} propose to decouple content and positional information in the decoder to provide better spatial priors in localization. 
\cite{liu2022dab, zhu2020deformable} further design better formulations of positional queries than previous works.
%
% Inspired by the denoising training in DN-DETR \cite{li2022dn},
The one-to-one label-assignment scheme is also widely discussed in \cite{li2022dn, zhang2022dino, jia2022detrs, chen2022group} for a better assignment.  
Moreover, some models design \cite{yao2021efficient, zhu2020deformable, zhang2022dino} better decoder query initialization by utilizing dense priors from the encoder.
% These works have achieved remarkable performance, for example, DINO \cite{zhang2022dino} achieved first place on COCO2017 object detection, outperforming classical detectors for the first time.
\\\textbf{Improving Multi-Scale Feature Extraction of DETR:}
\label{sec:related_efficient}
Though DETR-based models with multi-scale features have shown promising performance~\cite{zhu2020deformable,zhang2022dino}, especially for small object detection, their efficiency is still a concern for many applications.
In fact, multi-scale feature extraction has been widely studied in many CNN-based detectors for efficiency and effectiveness, such as FPN \cite{lin2017feature}, BiFPN \cite{tan2020efficientdet}, PANET \cite{liu2018path}, and NAS-FPN \cite{ghiasi2019fpn}, yet the efficiency of multi-scale DETR is under-explored.
%when using multi-scale encoded features that is important for small object detection. 
%
% The reason for this high cost is that all the multi-scale features will be processed by the dense encoder to produce contextualized features, which will be queried by the decoder to produce detection results. 
% % Another important direction is to design better Transformer encoder. 
% %
% Meanwhile, as self-attention in Transformers has quadratic complexity with token number, Vanilla DETR \cite{carion2020end} is not capable of using multi-scale features due to the prohibitive computational cost in the encoder. 
% %
% In DETR the number of feature tokens in a 4-scale feature pyramid is around $15000$, which increases around $20 \times$ compared to single scale feature. 
%
%To resolve this problem, 
Recently, a few works \cite{zhu2020deformable, zhang2022detr++, zhang2022towards, song2022extendable} have attempted to design efficient encoders. 

% However, these methods still largely lags behind detection performance compared to SOTA detectors.
%
Deformable DETR \cite{zhu2020deformable} proposes deformable attention, which can be used in the DETR encoder to sparsify the values in a self-attention layer by sampling only a few values for each query. The proposed deformable encoder leads to good detection results with an affordable computation cost, which has been widely acknowledged and applied in many vision tasks. However, compared with single-scale detectors, the computation cost of multi-scale deformable DETR is still high for efficient usage.
Based on the strong deformable encoder, some works attempt to improve its efficiency. Efficient DETR \cite{yao2021efficient} proposes to use \emph{fewer encoder layers} by leveraging encoder dense priors for decoder query initialization.
Sparse DETR \cite{roh2021sparse} proposes to sparsely update salient tokens in the encoder to \emph{reduce the number of queries} with a scoring network. In fact, the encoder is responsible for feature extraction, but Sparse DETR introduces multi-layer detection loss in encoder layers, making it hard to generalize to other DETR-based models. 

Recently, DETR++ \cite{zhang2022detr++} proposes to replace the encoder with BiFPN~\cite{tan2020efficientdet} and VIDT \cite{song2022extendable} develops a stronger decoder to remove the encoder. 
IMFA \cite{zhang2022towards} proposes to sample sparse scale-adaptive features from some interesting areas of multi-scale features. 
However, the performance of these models still largely lags behind improved detectors \cite{li2022dn, zhang2022dino} based on the deformable encoder. 

%% file: resources/tables/flopsmem.tex
\begin{table*}[h]
\begin{adjustbox}{width=0.85\textwidth,center}
\begin{minipage}[h]{1.\textwidth}
\makeatletter\def\@captype{table}
\centering
\begin{tabular}{c|c|ccc|c|ccc}
    \toprule
    Model &Total GFLOPs & Backbone  & (M.S.) Encoder& Decoder &Total Train Mem &AP&AP$_s$&AP$_L$  \\
        \midrule
        DINO-4scale ($100\%$) & 235 & 70 &\textbf{137}& 28  &32G &50.7&\textbf{33.5}&64.7  \\
        \hline
        DINO-3scale ($25\%$)& 122 & 70 &31& 21 &13G  & 48.2&\textbf{30.1}&63.9 \\
    
    \bottomrule
\end{tabular}
\end{minipage}\hspace{3mm}
\end{adjustbox}
\vspace{-.2cm}
\caption{GFLOPs of DINO based on ResNet-50 with four feature scales and three feature scales, respectively. We use ResNet-50 as the backbone and evaluate on COCO \emph{val2017} trained with 12 epochs. $100\%$ means we use all the feature tokens, while $25\%$ means we use three high-level features, which accounts for $25\%$ of all tokens.
}
\vspace{-.4cm}
\label{tab:dinoflops}
\end{table*}

%% file: resources/sec/method.tex
\input{resources/images/framework}
\section{Method}
\subsection{Motivation and Analysis}
\label{sec:preliminary}
%1. the basic formulation of DETR: (a)regard the detection as a set of object prediction problem; (b) backbone, encoder, decoder, and loss strategies to make an end-to-end detection); -> 引出遗留问题:1. 性能比传统的差 （后续引出一系列改进性能的工作，从机制上：主要改进decoder）；2.Transformer中的高开销（后续引出改进效率的工作，deformable detr -> multi-scale的重要性以及引入 -> 从开销大的部分：主要改进encoder）
In this part, we first analyze why existing DETR-based models are still inefficient and then show some interesting observations.
Multi-scale features are of vital importance for detecting  objects of diverse scales. 
%We try to understand how multi-scale feature works.
%Each feature scale is the output of the corresponding intermediate backbone stage. Normally, 
They are composed of multiple feature scales ranging from high-level (low resolution) to low-level (high resolution) features. Each lower-level feature map contains $4 \times$ more tokens than its previous feature level. 
From Table \ref{tab:feature scale}, we can observe that the number of tokens in low-level features quadratically increases, whereas the three higher-level scales account for only about $25\%$. 

\input{resources/tables/featurescale}

%
%Therefore, multi-scale features will increase the token number by $20 \times$, which is unacceptable for the encoder dense self-attention layers. A widely-adopted design of multi-scale encoder is the deformable encoder \cite{zhu2020deformable}, which sparsifies the values in attention to reduce complexity.  However, due to the large number of multi-scale tokens, the deformable encoder is still heavy for efficient use. 
Furthermore, we take a DETR variant DINO~\cite{zhang2022dino} as a preliminary example.
What will happen if we simply drop the low-level feature (S4 in Table \ref{tab:feature scale}) in its deformable encoder to reduce the computational costs?
In Table~\ref{tab:dinoflops}, a reduced DINO-3scale model trades a \textbf{48\%} efficiency gain in terms of GFLOPs at the cost of a \textbf{4.9\%} average precision (AP) and even \textbf{10.2\%} AP on small object detection deterioration. However, the AP on large objects is competitive.
%compare the efficiency and effectiveness of a standard DINO-4scale model and . %We can see its deformable encoder takes most of the computation and training memory.
%For example, the encoder takes around $83\%$ of the total GFLOPs in the SOTA detection model DINO \cite{zhang2022dino} that uses deformable encoder.
That is, high-level tokens contain compact information and rich semantics to detect most objects. By contrast, a large number of low-level tokens are mainly responsible for local details to detect small objects. 
Meanwhile, multi-scale features contain many redundant tokens, especially low-level features. Therefore, we would like to explore \emph{how to efficiently update multi-scale features by primarily focusing on constructing better high-level features.} 
%
%We ask: can we minimize the number of encoder tokens  while keep the performance effectively?
%Therefore, reducing the token number in encoder is a feasible way to alleviate computational cost.

% One natural idea to reduce encoder token number is to select the salient tokens in the dense multi-scale features, which is similar to Sparse DETR \cite{roh2021sparse}. 
% However, there are three drawbacks in this approach. First, it breaks the structured feature organization, which is hard to generalize in other other models or tasks. Secondly, with limited and implicit supervision on the scoring network, the selected tokens may not be the optimal for detection. Third, it introduces other components, such as multiple auxiliary encoder detection loss to enhance its sparse encoder representation.  However, encoder is responsible for feature extraction, adding detection supervision makes it is hard to generalize to existing models. In DINO \cite{zhang2022dino}, adding encoder detection loss will cause performance drop \footnote{In our experiments with a ResNet-50 backbone, adding encoder detection loss alone will cause 1.4 AP drop.}.

%However, the excessive number of low-level features contains mostly unnecessary local details. 
%As shown in Table \ref{tab:dinoflops}, using only 3 scale features will significantly reduce the encoder computational cost, while AP of large objects is comparable with the original 4-scale model. The performance drop mainly results from the small objects.
In this way, we can prioritize high-level feature updates in most layers, which could significantly reduce query tokens for a more efficient multi-scale encoder.
To sum up, this work aims to design a general solution for highly efficient DETR-based detectors and maintain competitive performance.

\subsection{Model Overview}
Following the multi-scale deformable DETR \cite{zhu2020deformable}, \modelname is composed of a backbone, a multi-layer encoder, and a multi-layer decoder with prediction heads. 
%The is based on the deformable attention, which only samples a few values for each query, while the encoder queries remains large for efficient use due to multi-scale features.
The overall model framework is shown in Fig. \ref{fig:framework}.
Specifically, 
we split the multi-scale features from a backbone into high-level features and low-level features. These features will be updated in an interleaved manner (introduced in Sec. \ref{sec:interleave}) with different updating frequencies (explained in Sec. \ref{sec:high} and \ref{sec:low}) in the proposed efficient encoder block to achieve precision and efficiency trade-off.
To enhance the lagged update of low-level features, we further introduce a key-aware deformable attention (KDA) approach ( described in Sec. \ref{sec:KDA}). 

\subsection{Interleaved Update}
\label{sec:interleave}
From our motivation, the bottleneck towards an efficient encoder is excessive low-level features, where most of which are not informative but contain local details for small objects. 
Moreover, multi-scale features $S$ are structured in nature, where the small number of high-level features encodes rich semantics but lack important local features
for some small objects. 
Therefore, we propose to prioritize different scales of the features in an interleaved manner to achieve a precision and efficiency trade-off.
We split ${S}$ into low-level features $F_L\in \mathbb{R}^{N_L\times d_{model}}$ and high-level features $F_H\in \mathbb{R}^{N_H\times d_{model}}$, where $d_{model}$ is the channel dimension, and $N_H$ and $N_L$ are the corresponding token number ($N_H \approx 6\%\sim33\% N_L$). $F_H$ can contain the first three or two scales in different settings, for clarity, we set $F_H$ to $S1, S2, S3$ and $F_L$ to $S4$ by default.
$F_H$ is regarded as the primary feature and is updated more frequently, whereas $F_L$ is updated less frequently. As deformable attention has a linear complexity with feature queries, the small number of frequently-updated high-level features largely reduces the computational cost.
As shown in Fig.~\ref{fig:framework}, we stack the efficient encoder block for $B$ times, where each block updates high-level features for $A$ times but only updates low-level features once at the end of the block. In this way, we can maintain a full-scale feature pyramid with a much lower computation cost. With this interleaved update, we design two effective updating mechanisms for $F_L$ and $F_H$.
%

% \subsection{High-level Feature Update}
\subsection{Iterative High-level Feature Cross-Scale Fusion}
\label{sec:high}
In this module, the high-level features $F_H$ will serve as queries ($\textbf{Q}$) to extract features from \emph{all-scales}, including the low-level and high-level feature tokens. This operation enhances the representation of $F_H$ with both high-level semantics and high-resolution details. 
The detailed updating process is shown in Fig. \ref{fig:framework}(a). 
This operation is highly efficient. For example, using multi-scale feature queries in the first two scales or the first three scales will significantly reduce $94.1\%$ and $75.3\%$ queries, respectively, as shown in Table \ref{tab:feature scale}. 
%
% The detailed updating process is shown in Fig. \ref{fig:framework}(a). 
We also use the proposed key-aware attention module {KDA}, which will be discussed in Sec \ref{sec:KDA}, to perform attention and update tokens. Formally, the update process can be described as
\begin{equation}
\begin{aligned}
\textbf{Q}&=F_H,\ \textbf{K}=\textbf{V}=Concat(F_H, F_L) \\
    F_H'&=KDA(\textbf{Q}, \textbf{K}, \textbf{V}) \\
    Output&=Concat(F_H', F_L) \\
\end{aligned}
\end{equation}
where $Concat$ is to concatenate low-level and high-level features into \emph{full-scale} features,  
query $\textbf{Q}$ is the initial high-level features, $\textbf{K}$ and $\textbf{V}$ are initial features from all levels, and $F_H$ is the high-level tokens, and
$F_H'$ are the updated high-level features.

A high-level feature update layer will be stacked for multiple (e.g., $A$ times) layers for iterative feature extraction. Note that the updated $F_H'$ will also update $\textbf{Q}$ and the corresponding high-level features in the multi-scale feature pyramid iteratively, which makes a feature update in $\textbf{K}$ and $\textbf{V}$ in the next layer.
% , as shown in the right part of Fig. \ref{fig:framework}(a). 
Interestingly, this high-level feature updating module is similar to the Transformer decoder, where we use a small number of high-level tokens to query their features similar to a self-attention and query a large number of low-level features similar to cross-attention. 
% The attention module used here is the proposed $KDA$ from Sec \ref{sec:KDA}. 
% We will stack it $M$ times in each efficient encoder block.

\subsection{Efficient  Low-level Feature Cross-Scale Fusion}\label{sec:low}
As shown in Table \ref{tab:feature scale}, the low-level features contain excessive tokens, which is a critical factor for inefficient computation. 
%
%Meanwhile, they are mainly responsible for local details for small objects. 
%, which is less important compared to high-level features. 
Therefore, the efficient encoder updates these low-level features at a lower frequency after a sequence of high-level feature fusion.
% After a sequence of reduced attention blocks, the updated reduced tokens encode both high-level semantics and important low-level details. To maintain structured feature organization of all scales, we use a feature extrication block to decode these low-level details in the reduced tokens back to the initial high-level features. 
Specifically, we utilize the initial low-level features as queries to interact with the updated high-level tokens as well as the original low-level features to update their representation. 
\input{resources/images/keyaware}
Similar to the high-level feature update, we use the interaction with a KDA attention layer. Formally, we have
\begin{equation}
\begin{aligned}
    \textbf{Q}&=F_L,\ \textbf{K}=\textbf{V}=Concat(F_H',F_L) \\
    F_L'&=KDA(\textbf{Q}, \textbf{K}, \textbf{V}) \\
    Output&=Concat(F_L', F_H')
\end{aligned}
\end{equation}
where $\textbf{Q}$ is from the original low-level features, $F_H'$ and $F_L'$ are the updated high-level and low-level features, respectively. After a KDA layer, we can obtain the $F_L'$. Finally, we construct the output multi-scale features $S'$ by concatenating the updated low-level and high-level features. 
To further reduce the computational cost, we use a lightweight feed-forward network with a hidden dimension size $\frac{1}{\lambda}$ of the original size. $\lambda$ is $8$ in our model. 
%The expanded low-level features are concatenated with the compressed tokens to construct the original multi-scale features.
\input{resources/tables/deformable}
\subsection{Key-aware Deformable Attention}\label{sec:KDA}
% \lf{Have rewritten related work and method part before this section}
In a vanilla deformable attention layer, the query $\textbf{Q}$ will be split into $M$ heads, and each head will sample $K$ points from each of the $L$ feature scales as value $\textbf{V}$. 
% $Q$ and $V$ are of $d_{model}$-dimensional vectors.
Therefore, the total number of values sampled for a query is $N_v=M\times L \times K$. The sampling offsets $\Delta p$, and their corresponding attention weights are directly predicted from queries using two linear projections denoted as $W^p$ and $W^A$. Deformable attention can be formulated as
\begin{equation}
\begin{aligned}
    % A &= QW^A \\
    \Delta &p = \textbf{Q}W^p, \textbf{V} = Samp(S, p+\Delta p)W^V \\
    D&eformAttn(\textbf{Q}, \textbf{V}) = Softmax(\textbf{Q}W^A)\textbf{V}
\end{aligned}
\end{equation}
% \begin{equation}
%     V^i_n=\left\{\begin{array}{ll}
%  m, &\text{if } O^i_n \text{ matches } T_m\\
%  -1, &\text{if } O^i_n \text{ matches nothing}
% \end{array}\right.
% \label{eq: matching}
% \end{equation}
Where the projections are parameter matrices $W^A, W^p \in \mathbb{R}^{d_{model}\times N_v}$ and $W^V \in \mathbb{R}^{d_{model}\times d_{model}}$. $p$ are the reference points of the query features and $\Delta p, p \in \mathbb{R}^{(N_H+N_L)\times N\times 2}$.  $S$ is the multi-scale feature pyramid. With the sampled offsets $\Delta p$, it computes the features with function $Samp(S, p+\Delta p)$ in the sampled locations $(p+\Delta p)$ of feature pyramid $S$ with bilinear interpolation. 
% After that, it sums the interpolated features $V$ with the projected weights as the output features.
% \begin{equation}
%     % V &= QW^V \\ 
%     DeformAttn(Q, V) = Softmax(A)V
% \end{equation}
% where $S$ is the total number of tokens sampled as values for each query.
Note that no key participates in the original deformable attention layer, indicating that a query can decide the importance of each sampled value by only its feature without comparing it with keys. 
%
% As there is a linear function to predict weights, the key can also be viewed as shared and learnable in the weights of linear functions. 
%In original deformable encoder, 
As all the multi-scale features will be the queries to sample locations and attention weights, the original model can quickly learn how to evaluate the importance of each sampled location given the queries. 
Nevertheless, the interleaved update in our encoder makes it difficult for the queries to decide both the attention weights and sampling locations in other asynchronous feature maps, as shown in Fig. \ref{fig:vis}.
\begin{figure}[t]
\vspace{-0.2cm}
    \centering
    \includegraphics[width=1\linewidth]{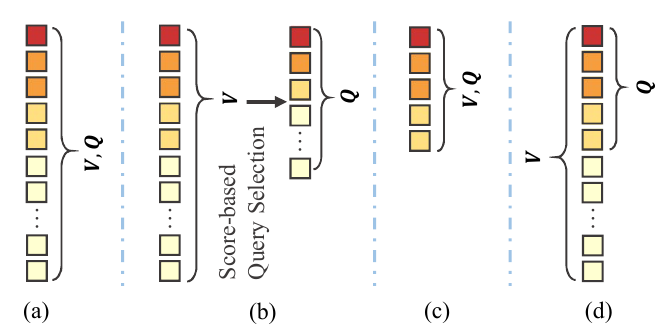}
    \caption{Comparison of previous efficient encoders strategies in (a) Deformable DETR~\cite{zhu2020deformable}, (b) Sparse DETR~\cite{roh2021sparse}, and (c) Trivially using only the first three high-level scales. (d) Preliminary efficient encoder to only update high-level features. We also present the results of (c) and (d) in Table \ref{tab:effective}.}
    \label{fig:compare_lite}
    \vspace{-0.4cm}
\end{figure}

To better fit the efficient encoder designs, we propose a key-aware deformable attention (KDA) approach to sampling both keys and values for a query, as shown in Fig. \ref{fig:keyaware}. The sampled keys and values, together with the query, will then perform a standard scaled dot-product attention. 
Formally, we have
\begin{equation}
\begin{aligned}
    % \Delta p &= QW^p \\
    \textbf{V} &= Samp(S, p+\Delta p)W^V, \\
    \textbf{K} &= Samp(S, p+\Delta p)W^K, \\
    KDA(\textbf{Q}, \textbf{K}, \textbf{V}) &= Softmax(\frac{\textbf{Q}\textbf{K}^T}{\sqrt{d_k}})\textbf{V}
    \end{aligned}
\end{equation}
where $d_k$ is the key dimension of a head. The computational complexity of $KDA$ is the same as the original deformable attention as we sample the same number of values for each query. In this way, KDA can predict more reliable attention weights when updating features from different scales.

\subsection{Discussion with Sparse DETR and other Efficient Variants}
Another efficient way is to reduce encoder tokens by selecting salient tokens in the multi-scale features, like Sparse DETR \cite{roh2021sparse}. 
However, there are three drawbacks to this kind of approach. First, it is hard to generalize across other DETR-based models since it breaks the structured feature organization.
Second, the selected tokens via a scoring network may not be optimal due to limited and implicit supervision. Third, it introduces other components, such as multiple auxiliary encoder detection loss, to enhance its sparse encoder representation. As the encoder is responsible for feature extraction, adding detection supervision makes it difficult to apply to existing models\footnote{In our experiments on DINO \cite{zhang2022dino} with a ResNet-50 backbone, adding the encoder detection loss alone will cause a \textbf{1.4} AP drop.}.

Moreover, we illustrate the previous efficient encoders and the preliminary efficient designs in Figure~\ref{fig:compare_lite} for a clear comparison.
\input{resources/tables/dino}

%% file: resources/images/framework.tex
\begin{figure*}[h]
%\vspace{-0.2cm}
    \centering
    \includegraphics[width=1\linewidth]{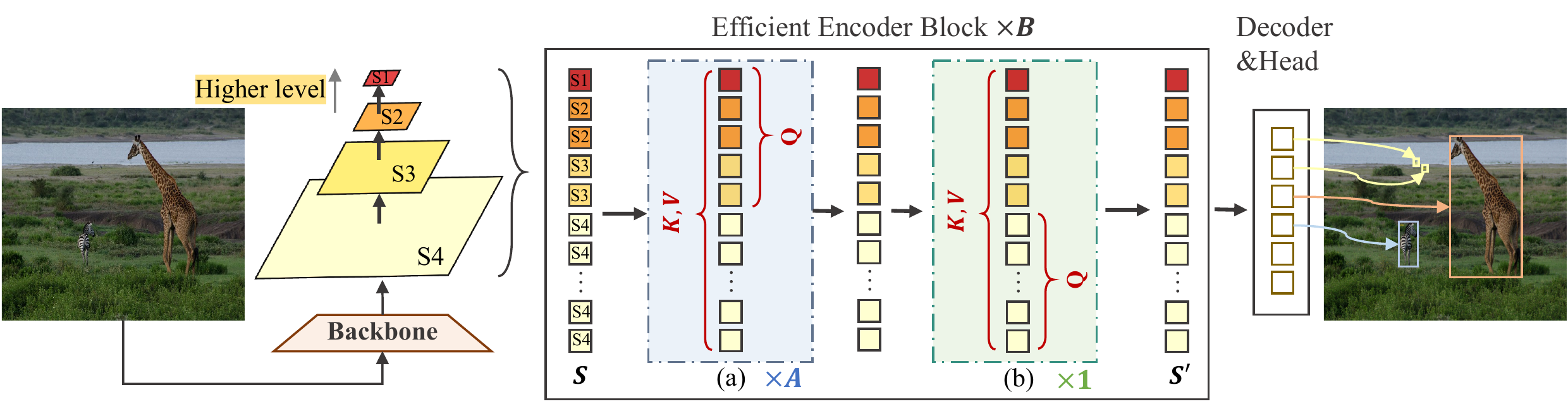}
    \caption{Illustration of the \modelname framework. We use $S_2\sim S_4$ to indicate the features from different backbone stages. That is, they correspond to $C_5\sim C3$ in ResNet-50~\cite{he2015deep}. $S_1$ is acquired by further downsampling $C_5$ by a ratio of 0.5. In this figure, we take $S1\sim S3$ as high-level features as an example. Moreover, (a) is the proposed high-level feature update discussed in Sec.~\ref{sec:high} and (b) is the low-level feature cross-scale fusion discussed in Sec.~\ref{sec:low}. In each efficient encoder block, the multi-scale features will go through high-level feature update for $A$ times and then conduct low-level feature update at the end of each block. The efficient encoder block will perform $B$ times. }
    \label{fig:framework}
    \vspace{-0.2cm}
\end{figure*}

%% file: resources/tables/featurescale.tex
\begin{table}[t]
\centering
\begin{adjustbox}{width=1.25\columnwidth}
\begin{minipage}[t]{1.3\columnwidth}
% \makeatletter\def\@captype{table}
\begin{tabular}{c|cccc}
    \toprule
    Feature Scale (S) &S1 ($\frac{1}{64}$) & S2 ($\frac{1}{32}$) & S3 ($\frac{1}{16}$)&S4 ($\frac{1}{8}$)  \\
        \midrule
        Token Ratio &1.17\% & 4.71\% & 18.8\%&75.3\%     \\
    \bottomrule
\end{tabular}
\end{minipage}\hspace{3mm}
\end{adjustbox}
\vspace{-.2cm}
\caption{The token ratio of each feature scale in a 4-scale feature pyramid. }
\vspace{-.4cm}
\label{tab:feature scale}
\end{table}

%% file: resources/images/keyaware.tex
\begin{figure}[t]
% \vspace{-0.8cm}
    \centering
    \includegraphics[width=0.5\textwidth]{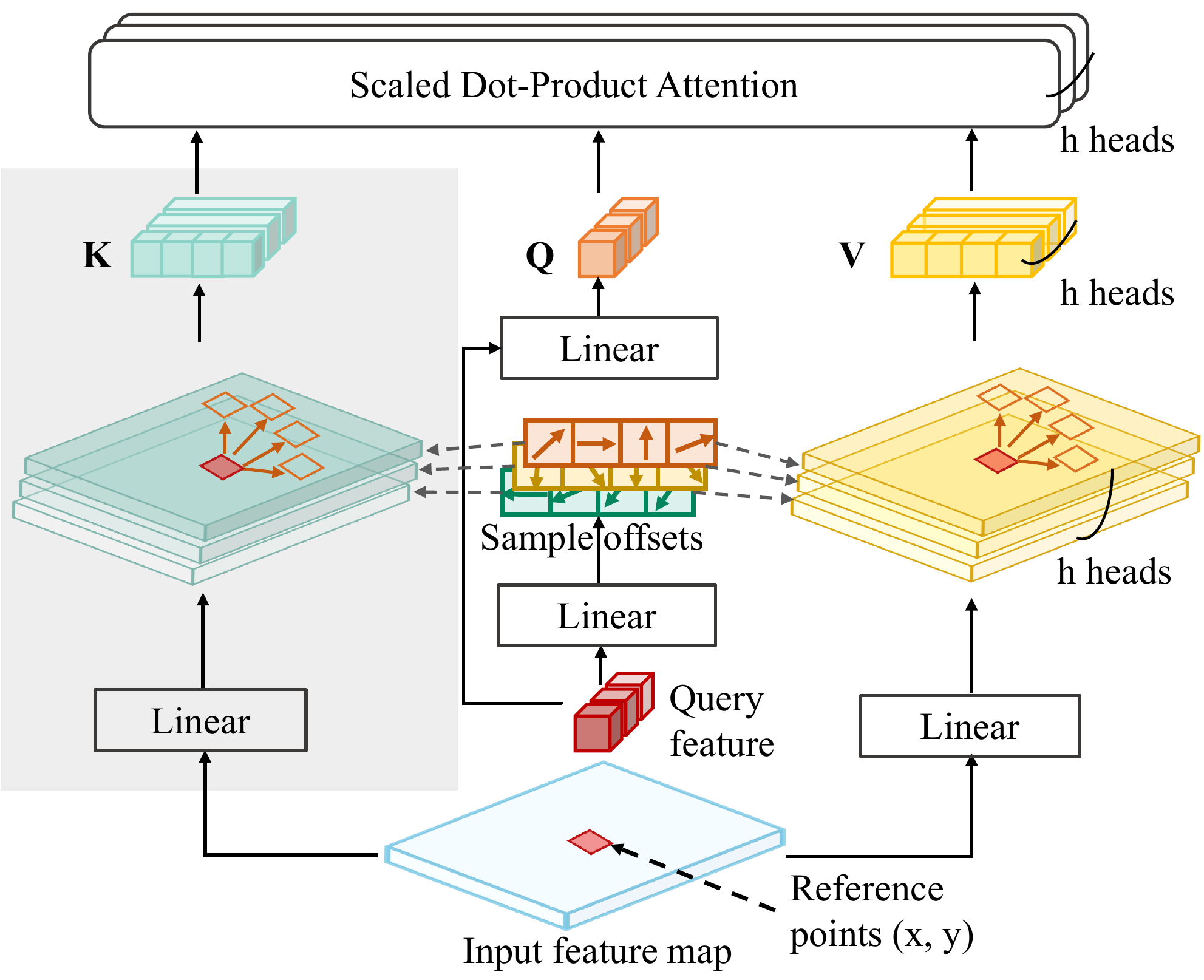}
    % \vspace{-0.5cm}
    \caption{Illustration of the proposed Key-aware Deformable Attention (KDA) layer. }
    \label{fig:keyaware}
    \vspace{-0.4cm}
\end{figure}

%% file: resources/tables/deformable.tex
\begin{table*}[t]
    \centering
   \resizebox{.9\textwidth}{!}{%
    \begin{tabular}{lcccccccc>{\columncolor{mygray} }cc}
        \toprule
        Model & \#epochs & AP & AP$_{50}$ & AP$_{75}$ & AP$_{S}$ & AP$_{M}$ & AP$_{L}$ & GFLOPs&\begin{tabular}[l]{@{}l@{}}Encoder \\ GFLOPs
        \end{tabular} & Params \\
        \midrule
        % DETR~\cite{carion2020end}                & $500$ & $42.0$ & $62.4$ & $44.2$ & $20.5$ & $45.8$ & $61.1$ & $86$ & $41$M \\
        % Faster RCNN-FPN~\cite{ren2015faster}   & $108$ & $42.0$  & $62.1$ & $45.5$ & $26.6$ & $45.5$ & $53.4$ & $180$ & $42$M \\
        % Anchor DETR~\cite{wang2021anchor}         & $50$ & $42.1$ & $63.1$ &  $44.9$ &  $22.3$ &  $46.2$  & $60.0$ & $-$ & $39$M \\
        % Conditional DETR~\cite{meng2021conditional}    & $50$ & $40.9$ & $61.8$ & $43.3$ & $20.8$ & $44.6$ & $59.2$ & $90$ & $44$M \\
        % DAB-DETR~\cite{liu2022dab}    & $50$ & $42.2$ & $63.1$ & $44.7$ & $21.5$ & $45.7$ & $60.3$ & $94$ & $44$M \\
        % DN-DETR& $50$ & $44.1$ & $64.4$ & $46.7$ & $22.9$ & $48.0$ & $63.4$ & $94$ & $44$M \\
        
         DETR-DC5~\cite{carion2020end}               & $500$ & $43.3$ & $63.1$ & $45.9$ & $22.5$ & $47.3$ & $61.1$ & $187$&$~100$ & $41$M \\
        Anchor DETR-DC5~\cite{wang2021anchor}           & $50$ & $44.2$ & $64.7$ & $47.5$ & $24.7$ & $48.2$ & $60.6$ & $151$&$~70$ & $39$M \\
        Conditional DETR-DC5~\cite{meng2021conditional}  & $50$  &  $43.8$ & $64.4$ &  $46.7$ &  $24.0$ & $47.6$ &  $60.7$ & $195$&$~100$ & $44$M \\
        DAB-DETR-DC5~\cite{liu2022dab}    & $50$ & {${44.5}$} & $65.1$ & $47.7$ & $25.3$ & $48.2$ & $62.3$ & $202$ &$~100$& $44$M \\
        DN-DETR-DC5    & $50$ & ${46.3}$ & $66.4$ & $49.7$ & $26.7$ & $50.0$ & $64.3$ & $202$ &$~100$& $44$M \\
        \hline
        \hline
        \textbf{Deformable DETR efficient variants}\\
        \hline
        % Deformable DETR$^\dag$ \cite{zhu2020deformable}&$50$&46.8&65.2&50.0&28.8&49.2&61.7&177&$90$&40M \\
        Deformable DETR$^\dag$ \cite{zhu2020deformable}&$50$&46.8&66.0&50.6&29.8&49.7&62.0&177&$90$&40M \\
        \hline
        \prefix-Deformable DETR H2L2-(2+1)x3(5\%, ours)&$50$&45.8&65.1&49.3&27.7&49.1&61.1&108&\textbf{23($\downarrow$ 74\%)}&41M \\
        \prefix-Deformable DETR H3L1-(6+1)x1(25\%, ours)&$50$&45.9&65.6&49.2&27.9&49.0&61.6&115&\textbf{30($\downarrow$ 66\%)}&41M \\
        \prefix-Deformable DETR H3L1-(3+1)x2(25\%, ours)&$50$&46.2&65.5&49.8&28.2&49.2&61.5&119&\textbf{35($\downarrow$ 61\%)}&41M \\
        \prefix-Deformable DETR H3L1-(2+1)x3(25\%, ours)&$50$&\textbf{46.7}&66.1&50.6&29.1&49.7&62.2&123&\textbf{39($\downarrow$ 57\%)}&41M \\
        \hline
        Efficient DETR \cite{yao2021efficient}&$50$&44.2&62.2&48.0&28.4&47.5&56.6&159&$~79$&32M \\
        \hline
        Sparse DETR$^*$-rho-0.1 \cite{zhu2020deformable}&$50$&45.3&65.8&49.3&28.4&48.3&60.1&111&$24$&41M \\
        Sparse DETR$^*$-rho-0.2 \cite{zhu2020deformable}&$50$&45.6&65.8&49.6&28.5&48.6&60.4&119&$32$&41M \\
        Sparse DETR$^*$-rho-0.3 \cite{zhu2020deformable}&$50$&46.0&65.9&49.7&29.1&49.1&60.6&127&$40$&41M \\
        Sparse DETR$^*$-rho-0.5 \cite{zhu2020deformable}&$50$&46.3&66.0&50.1&29.0&49.5&60.8&141&$54$&41M \\

        \hline
        \bottomrule
    \end{tabular}}
    \centering
    \vspace{-.2cm}
    \caption{Results for single-scale DETR-based models which use a larger resolution feature map with dilation (DC5) and Deformable DETR-based models for improving efficiency. All models are based on ResNet-50. 
    % 'DC5' use a dilated larger resolution feature map.  
    $^*$ Sparse DETR is based on an improved Deformable DETR baseline that combines the components from Efficient DETR \cite{yao2021efficient}. 'rho' is the keeping ratio of encoder tokens in Sparse DETR. Value in the parenthesis indicates the percentage of our high-level tokens compared to the original features. $^\dag$ we adopt the result from the official Deformable DETR codebase. The meaning of different model variants is described in Sec. \ref{setup}.
    % The models with superscript $^{*}$ use $3$ pattern embeddings as in Anchor DETR~\cite{wang2021anchor}.
    }
    \label{tab:deformbale}
    % \vspace{-.3cm}
\end{table*}

%% file: resources/tables/dino.tex
% \rowcolors{1}{blue}
\begin{table*}[t]
    \centering
   \resizebox{0.9\textwidth}{!}{%
    \begin{tabular}{ccccccccc >{\columncolor{mygray} }cccc}
        \toprule
        Model & \#epochs & AP & AP$_{36}$ & AP$_{75}$ & AP$_{S}$ & AP$_{M}$ & AP$_{L}$ & GFLOPs &\begin{tabular}[l]{@{}l@{}}Encoder \\ GFLOPs
        \end{tabular}& Params \\
        \midrule
        EfficientDet-D6 \cite{tan2020efficientdet}&$-$&51.3&$-$&$-$&$-$&$-$&$-$&226&$-$&52M\\
        YOLOv5-X \cite{yolov5}&$-$&50.7&$-$&$-$&$-$&$-$&$-$&206&$-$&87M\\
         YOLOv7-X \cite{wang2022yolov7}&$-$&52.9&$-$&$-$&$-$&$-$&$-$&190&$-$&71M\\
        %  YOLOv5-X \cite{glenn_jocher_2020_4154370}&-&36.7&&&&&&206&$-$&87M\\
        %  YOLOv7-X \cite{wang2022yolov7}&-&52.9&&&&&&190&$-$&71M\\
        \hline
        \hline
        % \multicolumn{5}{c}{{\textbf{Swin-T backbone}}}\\
        \textbf{Swin-T backbone} \\
        \hline
        VIDT+ \cite{song2022extendable}&50&49.7&67.7&54.2&31.6&53.4&65.9&$-$&$-$&38M\\
        D$^2$ETR \cite{lin2022d}&50&49.1&$-$&$-$&$-$&$-$&$-$&$127$&$-$&46M\\
         \midrule[1pt]
        DINO \cite{zhang2022dino}&$36$&54.1&72.0&59.3&38.3&57.3&68.6&243&$137$&47M \\

        \hline
        % \prefix-DINO C2-6x1(5\%, ours)&$36$&$-$&-.4&-.9&-.6&-.0&-.8&124&$\textbf{18(13\%)}$&47M \\
        \prefix-DINO H2L2-(2+1)x3(5\%, ours)&$36$&$53.1$&71.4&57.9&36.6&56.0&68.8&138&\textbf{30($\downarrow$78\%)}&47M \\
        \prefix-DINO H3L1-(6+1)x1(25\%, ours)&$36$&$53.3$&71.7&58.2&36.3&56.6&68.7&149&\textbf{41($\downarrow$70\%)}&47M \\
        % Efficient-DINO C3-3x2(ours)&$36$&46.2&&&&&&119&40 \\
        \prefix-DINO H3L1-(2+1)x3(25\%, ours)&$36$&$\textbf{53.9}$&72.0&58.8&37.9&57.0&69.1&159&\textbf{53($\downarrow$62\%)}&47M \\
        
        \midrule[1pt]
        H-DETR \cite{jia2022detrs}&$36$&53.2&71.5&58.2&35.9&56.4&68.2&234&$137$&47M \\
        \hline
        % \prefix-DINO C2-3x2(ours)&$36$&-&&&&&&-&47M \\
        % \prefix-DINO H3L1-(6+1)x1(ours)&$36$&-&&&&&&139&$41$&47M \\
        % Efficient-DINO C3-3x2(ours)&$36$&46.2&&&&&&119&40 \\
        \prefix-H-DETR H2L2-(2+1)x3(5\%, ours)&$36$&52.3&70.7&57.2&35.9&55.2&67.7&131&$30$&47M \\
        \prefix-H-DETR H3L1-(6+1)x1(25\%, ours)&$36$&52.7&71.5&58.3&35.6&56.0&68.0&142&$41$&47M \\
        \prefix-H-DETR H3L1-(2+1)x3(25\%, ours)&$36$&\textbf{53.0}&71.3&58.2&36.3&56.3&68.1&152&$53$&47M \\
         \hline
         \hline
        \textbf{ResNet-50 backbone}\\
         \hline
         DFFT \cite{chen2022efficient}&36&$46.0$&$-$&$-$&$-$&$-$&$-$&101&$18$&$-$\\
         PnP-DETR \cite{wang2021pnp}&36&$43.1$&$63.4$&$45.3$&$22.7$&$46.5$&$61.1$&104&$29$&$-$\\
         AdaMixer \cite{gao2022adamixer} &$36$&47.0&66.0&51.1&30.1&50.2&61.8&132&$-$&135M\\
         IMFA-DETR \cite{zhang2022towards} &$36$&45.5&45.0&49.3&27.3&48.3&61.6&108&$\approx20$&53M\\
         \midrule[1pt]
        DINO \cite{zhang2022dino}&$36$&50.7&68.6&55.4&33.5&54.0&64.8&235&$137$&47M \\
        \hline
        % \prefix-DINO C2-3x2(ours)&$36$&-&&&&&&-&47M \\
        \prefix-DINO H2L2-(2+1)x3 (ours)&$36$&49.9&68.2&54.6&32.3&52.9&64.7&130&$30$&47M \\
        \prefix-DINO H3L1-(6+1)x1(ours)&$36$&50.2&68.6&54.3&33.0&53.4&66.0&141&$41$&47M \\
        % Efficient-DINO C3-3x2(ours)&$36$&46.2&&&&&&119&40 \\
        \prefix-DINO H3L1-(2+1)x3(ours)&$36$&\textbf{50.4}&68.5&54.6&33.5&53.6&65.5&151&$53$&47M \\
         \midrule[1pt]
        H-DETR \cite{jia2022detrs}&$36$&50.0&68.3&54.4&32.9&52.7&65.3&226&$137$&47M \\
        \hline
        \prefix-H-DETR H3L1-(2+1)x3 (ours)&$36$&49.5&67.6&53.9&32.0&52.8&64.0&142&$53$&47M \\
        \bottomrule
    \end{tabular}}
    \centering
    \vspace{-.2cm}
    \caption{Results for Deformable DETR-based models to improve efficiency with our light encoder design. We also compare with some efficient CNN-based models and other efficient DETR-based models. All models except EfficientDet and YOLO series are based on ResNet-50 and Swin-T pre-trained on ImageNet-1K. Percentage in the model name indicates the percentage of our compressed tokens compared to the original features. The meaning of different model variants is described in Sec. \ref{setup}.
    % The models with superscript $^{*}$ use $3$ pattern embeddings as in Anchor DETR~\cite{wang2021anchor}.
    }
    \label{tab:dino}
    \vspace{-.3cm}
\end{table*}

%% file: resources/sec/exp.tex
\section{Experiments}
\subsection{Setup}\label{setup}
We demonstrate the generalization capability of our proposed efficient encoder on a series of DETR-based models. We also evaluate the effectiveness of each component with ablations.\\
\textbf{Datasets: } We study \modelname on the challenging MS COCO 2017 \cite{lin2015microsoft} detection dataset. Following the common practice, we train on the training split and  report the detection performance on the validation split \textbf{val2017}. 
% For the evaluation metric, 
We report the standard mean average precision (AP) result under different IoU thresholds and object scales.
\\\textbf{Implementation details: } We evaluate the performance of \modelname  on multiple DETR-based models, including Deformable DETR \cite{zhu2020deformable}, H-DETR \cite{jia2022detrs}, and DINO \cite{zhang2022dino}. These models share a similar structure that is composed of a backbone, a multi-layer Transformer encoder, and a multi-layer Transformer decoder. Therefore, we simply replace their encoder with our proposed efficient module. Other model components are kept the same as the original model. In our KDA attention, for a fair comparison, we follow deformable attention to use M=8 and K=4.

We train the model for $36/50$ epochs according to the settings of different baselines and drop the learning rate in the $30/40$th epoch by a factor of $0.1$, respectively. Other model parameters, such as query number and hidden dimension size, are kept the same as in the original models. We use two backbones ResNet-50 \cite{he2015deep} and Swin-T \cite{liu2021swin} pre-trained on the ImageNet-1K \cite{deng2009imagenet} dataset in our experiments.

Following the common practice, we use augmentation in DETR \cite{carion2020end} to resize images so that the short side is between $480$ and $800$ while the long side is at most $1333$ pixels, followed by a random crop with probability $0.5$.
\\\textbf{Efficient encoder variants: }In our proposed efficient encoder block, three hyperparameters control the computational cost, including the number of high-level feature scales $H$ used in $F_H$, the number of efficient encoder blocks $B$, and the number of iterative high-level feature cross-scale fusion $A$. Therefore, we use $HL$-$(A+1)\times B$ to denote each variant of our \modelname, where $L$ is the number of low-level feature scales, and +1 denotes the default efficient low-level cross-scale feature fusion at the end of each block
% and 1 denote the efficient low-level feature cross-scale fusion in each efficient encoder block
. 
For example, \prefix-DINO H3L1-(3+1)$\times$2 indicates we base on DINO to use three high-level feature scales (H3L1) and two efficient encoder blocks with three high-level fusion ((3+1)$\times$2).

\subsection{Efficiency Improvements on Deformable DETR}
In Table \ref{tab:deformbale}, we use our proposed lite encoder to replace the deformable encoder in Deformable DETR and build \prefix-Deformable DETR. We achieve comparable performance as Deformable DETR with around $40\%$ of the original encoder GFLOPs. We can also observe that DETR-based models with a single scale of larger feature maps are computationally inefficient and inferior to multi-scale models.
In iterative high-level cross-scale fusion, we can effectively adopt high-level maps with only two or three high-level maps, which can reduce the queries in an encoder layer to $5\%\sim25\%$ of the original tokens. Compared with other efficient variants based on Deformable DETR, we achieve better performance under the same computational cost. For example, we outperform Sparse DETR-rho-0.3 by $0.7$ AP with fewer GFLOPs. In addition, Sparse DETR is based on an improved baseline that combines Efficient DETR and Deformable DETR. By contrast, our \prefix-Deformable DETR is simple and effective.
\input{resources/images/visualization}
\subsection{Efficiency Improvements on Other DETR-based Models}
Compared with other efficient variants, our efficient design is not constrained to  a specific detection framework and can be easily plugged into other DETR-based models. We take DINO \cite{zhang2022dino} and H-DETR \cite{jia2022detrs} as examples to show the effectiveness of our efficient encoder. The results are shown in Table \ref{tab:dino}. Compared with other recently proposed efficient DETR-like models \cite{gao2022adamixer, zhang2022towards}, our model achieves significantly better performance with comparable computational cost. In addition, after plugging in our efficient encoder, the encoder GFLOPs can be reduced by $78\%\sim62\%$ compared to the original ones while keeping $99\%$ of the original performance. Specifically, based on Swin-Tiny, our \prefix-DINO achieves \textbf{53.9} AP  with only $159$ GFLOPs, which also outperforms YOLO series models~\cite{wang2022yolov7, yolov5} under the same GFLOPs.
\subsection{Visualization of KDA } We also provide the visualization of our KDA attention in our interleaved encoder in Fig.~\ref{fig:vis}. Compared with deformable attention, as we introduce keys, our KDA attention can predict more reliable weights, especially on low-level feature maps. For example, in Fig. \ref{fig:vis}(a), the sampled locations of deformable attention in S4 (denoted with triangles) are less reliable compared to KDA. In Fig.~\ref{fig:vis}(b) and (c), we observe that it is difficult for deformable attention to focus on  meaningful regions on the largest scale map S4 in our interleaved encoder. KDA effectively mitigates this phenomenon, which helps extract better local features to bring the performance of small objects back.
\input{resources/tables/effectiveness_component}
\input{resources/tables/block_num}

\subsection{Ablation Studies}
% \\\textbf{Influence of different block number: }
\noindent\textbf{Effectiveness of each proposed component. } In Table \ref{tab:effective}, we show the effectiveness of our proposed components. We choose DINO-3scale and DINO-2scale as our baseline, which only uses the first three and two high-level feature maps. The results indicate that each of our proposed components requires a small computational cost while improving the model performance by a decent margin. Specifically, these components effectively bring the performance on small objects back, for example, the $AP_s$ of our efficient DINO-3scale is comparable with the original DINO-4scale model.
% In Row 1, we add the \textit{feature compression} and \textit{compressed attention} to the model, which builds compressed high-level tokens that fuses the low-level features while keep only 3 features scale in the encoder. The performance improvement indicates the compressed tokens contains local details to improve the detection performance. In Row 2, we additionally adds the \textit{feature extrication} to decode the local details back from the low-level features and construct a feature pyramid with 4 scales. This operation further improves the performance, which shows that structured multi-scale feature maps with more low-level features is important. Finally, we adds the proposed KDA attention, which shows +$0.6$ performance improvement.
\\\textbf{Influence of stacking the different number of modules. } In Table \ref{tab:block_num}, we explore the optimal choice to stack each module in our proposed efficient block.  
Based on Deformable DETR \cite{zhu2020deformable} with a ResNet-50 backbone,
we vary three arguments that influence the computational complexity and detection performance, including the number of high-level scales $H$ used as high-level features, efficient encoder block $B$, and iterative high-level feature cross-scale fusion $A$. 
% Except for the last row, we keep the total number of blocks the same as the original encoder with 6 layers.
The performance improves when we use more high-level feature scales and more encoder blocks to update the low-level features. However, further increasing the module number to $(2+1)\times 4$ will not improve the performance.

%% file: resources/images/visualization.tex
% \begin{figure}[htbp]
% \centering
% % \begin{subfigure}[]{.5\textwidth}
% \begin{minipage}[h]{0.24\linewidth}
% % \centering
% \includegraphics[width=1.3in]{resources/images/0_deform_l2.png}
% \subcaption{}
% \end{minipage}%
% \begin{minipage}[h]{0.24\linewidth}
% % \centering
% \includegraphics[width=1.3in]{resources/images/0_deform_l3.png}
% \subcaption{}
% \end{minipage}%
% % }%
% % \subfigure[b]{
% \begin{minipage}[h]{0.24\linewidth}
% % \centering
% \includegraphics[width=1.3in]{resources/images/1_deform_l2.png}
% \subcaption{}
% \end{minipage}
% % }
% % \subfigure[c]{
% \begin{minipage}[h]{0.24\linewidth}
% % \centering
% \includegraphics[width=1.3in]{resources/images/1_deform_l3.png}
% \subcaption{}
% \end{minipage}
% % % }
% % % \subfigure[d]{
% \begin{minipage}[h]{0.24\linewidth}
% % \centering
% \includegraphics[width=1.3in]{resources/images/0_kda_l2.png}
% \subcaption{}
% \end{minipage}
% % % }
% % % \subfigure[e]{
% \begin{minipage}[h]{0.24\linewidth}
% % \centering
% \includegraphics[width=1.3in]{resources/images/0_kda_l3.png}
% \subcaption{}
% \end{minipage}
% % % }
% % % \subfigure[f]{
% \begin{minipage}[h]{0.24\linewidth}
% \includegraphics[width=1.3in]{resources/images/1_kda_l2.png}
% \subcaption{}
% \end{minipage}
% \begin{minipage}[h]{0.24\linewidth}
% \includegraphics[width=1.3in]{resources/images/1_kda_l3.png}
% \subcaption{}
% \end{minipage}

% % \centering
% \caption{visualization of KDA attention. The upper row is the attention map from defomable attention, the }
% \label{fig:analyze_2st}
% \end{figure}
\begin{figure*}[h]
\vspace{-0.3cm}
    \centering
    \includegraphics[width=1\linewidth]{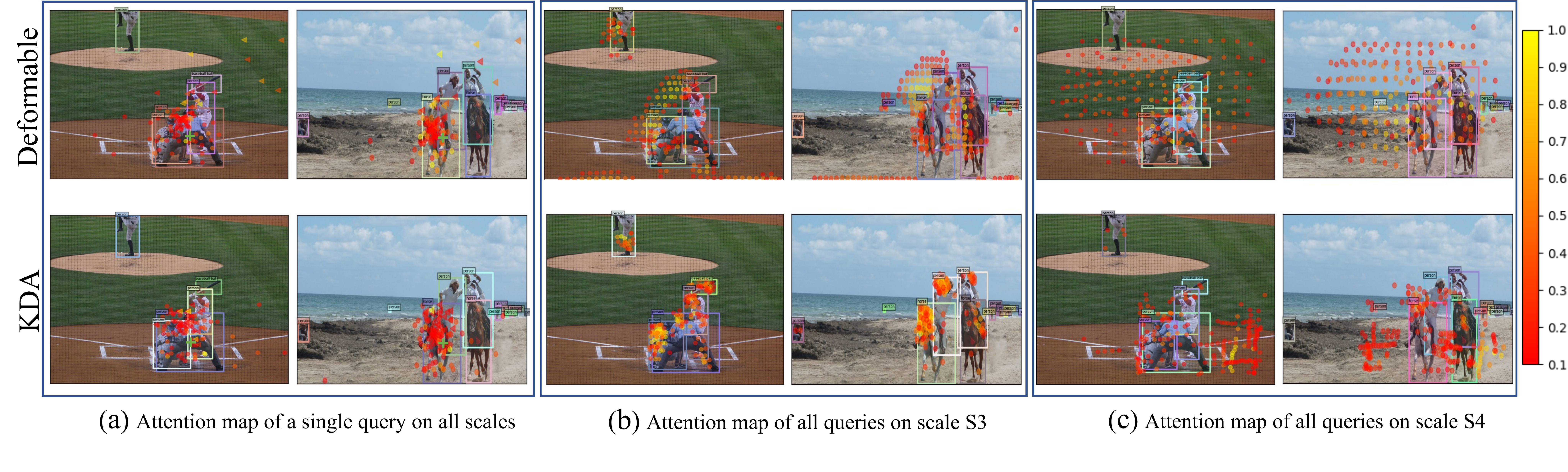}
    \vspace{-0.9cm}
    \caption{Visualization of KDA attention in our interleaved encoder. The first and second row are the attention maps of using deformable and our KDA attention. (a) We use the the center of an object from S1 (marked with "+" in green) as query and draw top 100 sampling locations on all four scales according to their attention weights. The sampling locations on S4 are marked with a triangle shape. (b)\&(c) We show top 200 sampling locations on scale S3 (b) and S4 (c) for all query tokens. The visualization shows that KDA can produce more reliable attention weights on high-resolution maps. For clarity, we only draw the locations of top 200 attention weights out of all sampling locations ($N_q\times M \times K$, $N_q$ is the total number of multi-scale query tokens) on S3 and on S4. More visualizations are shown in Appendix.}
    \label{fig:vis}
    \vspace{-0.4cm}
\end{figure*}

%% file: resources/tables/effectiveness_component.tex
\begin{table}[h]
\centering
\resizebox{0.45\textwidth}{!}{%
\begin{tabular}{c|ccc|ccccc}
    \toprule
    &\textbf{HL} & \textbf{LL} &\textbf{KDA}& \textbf{AP $\uparrow$}&\textbf{$AP_s$ $\uparrow$}& \textbf{GFLOPs $\downarrow$}&\begin{tabular}[l]{@{}l@{}}\textbf{Encoder} \\ \textbf{GFLOPs}
        \end{tabular}$\downarrow$  \\
    \midrule
    \multicolumn{4}{c|}{\textbf{DINO-4scale}\cite{zhang2022dino}} & 50.7&33.5& 235&137\\
    % \multicolumn{3}{c|}{\textbf{DINO-3scale}} & 48.2& 121&23\\
    \midrule
    % \multicolumn{3}{c|}{\textbf{DINO-3scale}}\\
    \multirow{5}{*}{{\rotatebox{90}{\textbf{3scale}}}}&
     $-$&$-$&$-$&48.2&30.1 &122 &31\\
    &$-$&$-$&\checkmark& 49.0\fontsize{8.0pt}{\baselineskip}\selectfont{(\textbf{+0.8})}&31.5&125&34\\
    &\checkmark &  $-$&$-$&49.0\fontsize{8.0pt}{\baselineskip}\selectfont{(\textbf{+0.8})}& 31.1& 128&  37\\
    &\checkmark&\checkmark& $-$& 49.8\fontsize{8.0pt}{\baselineskip}\selectfont{(\textbf{+0.8})}&33.0&147& 49\\
    &\checkmark&\checkmark& \checkmark&                                                                50.4\fontsize{8.0pt}{\baselineskip}\selectfont{(\textbf{+0.6})}&\textbf{33.5}&151&53\\
    \toprule
    % \multicolumn{3}{c|}{\textbf{DINO-2scale}}\\
    \multirow{3}{*}{{\rotatebox{90}{\textbf{2scale}}}}&
    $-$&$-$&$-$&45.2&24.1&113&14\\
    &$\checkmark$&$\checkmark$&$-$&49.2\fontsize{8.0pt}{\baselineskip}\selectfont{(\textbf{+4.0})}&31.8&$126$&26\\
    &$\checkmark$&$\checkmark$&$\checkmark$&49.9\fontsize{8.0pt}{\baselineskip}\selectfont{(\textbf{+0.7})}&\textbf{32.3}&130&30\\
    % \multicolumn{3}{c|}{\textbf{DINO-3scale}} & 48.2& 121&23\\
    \bottomrule
\end{tabular}
}
\caption{Effectiveness of each component on COCO \textit{val2017} trained with 36 epochs. The results are based on DINO with a ResNet-50 backbone trained for 36 epochs. HL means iterative high-level feature cross-scale fusion, LL means efficient low-level feature cross-scale fusion, and KDA is key-aware deformable attention.}
\vspace{-.3cm}
\label{tab:effective}
\end{table}

%% file: resources/tables/block_num.tex
\begin{table}[t]
    \centering
   \resizebox{0.5\textwidth}{!}{%
    \begin{tabular}{lccccc}
        \toprule
        \textbf{Model}  & \textbf{AP}$\uparrow$ &\textbf{$AP_s$ $\uparrow$}&\begin{tabular}[l]{@{}l@{}}\textbf{Encoder} \\ \textbf{GFLOPs}
        \end{tabular} $\downarrow$ \\
        \midrule
        Deformable DETR-4scale \cite{zhu2020deformable}&46.8&29.8&$90$ \\
        \hline
        Deformable DETR-2scale&40.3&20.4&$9$ \\
        \prefix-Deformable DETR H2L2-(2+1)x3 (ours)&45.8\fontsize{8.0pt}{\baselineskip}\selectfont{(\textbf{+5.5})}&27.7\fontsize{8.0pt}{\baselineskip}\selectfont{(\textbf{+7.3})}&$23$ \\
        \hline
        Deformable DETR-3scale&44.0&26.6&$16$ \\
        \prefix-Deformable DETR H2L2-(6+1)x1 (ours)&45.9&27.9&$28$ \\
        \prefix-Deformable DETR H3L1-(3+1)x2(ours)&46.2&28.2&$32$ \\
        \prefix-Deformable DETR H3L1-(2+1)x3(ours)&\textbf{46.7\fontsize{8.0pt}{\baselineskip}\selectfont{(\textbf{+2.7})}}&\textbf{29.1\fontsize{8.0pt}{\baselineskip}\selectfont{(\textbf{+2.5})}}&$36$ \\
        \prefix-Deformable DETR H3L1-(2+1)x4(ours)&46.6&29.6&$50$ \\
        \bottomrule
    \end{tabular}}
    \centering
    \vspace{-.3cm}
    \caption{Ablation study on stacking different number of each module in our efficient encoder block. All the models are built upon Deformable DETR-ResNet50 and trained for 50 epochs.
    }
    \label{tab:block_num}
    \vspace{-.3cm}
\end{table}

%% file: resources/sec/conclusion.tex
\section{Conclusion}
In this paper, we have analyzed that multi-scale features with excessive low-level features in the Transformer encoder are the primary cause of the inefficient computation in DETR-based models. We have presented \modelname with an efficient encoder block, which splits the encoder tokens into high-level and low-level features. These features will be updated in different frequency with cross-scale fusion to achieve precision and efficiency trade-off. To mitigate the effects of asynchronous feature, we further proposed a key-aware deformable attention, which effectively brings the detection performance of small objects back. As a result, our proposed efficient encoder can reduce computational cost by $60\%$ while keeping $99\%$ of the original performance. In addition, this efficient design can be easily plugged into many DETR-based detection models. We hope \modelname can provide a simple baseline for efficient detection in DETR-based models to benefit other resource-constrained applications.
\\\textbf{Limitations: }In this paper we mainly focus on reducing the computational complexity and do not optimize the run-time implementation of DETR-based model. We leave this to our future work.

%% file: resources/tables/sparselargecompare.tex
\begin{table*}[htb]
    \centering
   \resizebox{.9\textwidth}{!}{%
    \begin{tabular}{lcccccc>{\columncolor{mygray}}cccc}
        \toprule
        Model & \#epochs & AP & AP$_{50}$ & AP$_{75}$ & AP$_{S}$ & AP$_{M}$ & AP$_{L}$ & GFLOPs&\begin{tabular}[l]{@{}l@{}}Encoder \\ GFLOPs
        \end{tabular} & Params \\
        \midrule
        Deformable DETR$^\dag$ \cite{zhu2020deformable}&$50$&46.8&66.0&50.6&29.8&49.7&62.0&177&$90$&40M \\
        \hline
        \prefix-Deformable DETR H3L1-(2+1)x3(25\%, ours)&$50$&\textbf{46.7}&66.1&50.6&29.1&49.7&\textbf{62.2}&123&{39}&41M \\
        \hline
        Sparse DETR$^*$-rho-0.3 \cite{zhu2020deformable}&$50$&46.0&65.9&49.7&29.1&49.1&60.6&127&$40$&41M \\
        \hline
        \bottomrule
    \end{tabular}}
    \centering
    % \vspace{-.2cm}
    \caption{Results for Sparse DETR and Lite-Deformable DETR under the ResNet-50 backbone. 
    % 'DC5' use a dilated larger resolution feature map.  
    $^*$ Sparse DETR is based on an improved Deformable DETR baseline that combines the components from Efficient DETR \cite{yao2021efficient}. 'rho' is the keeping ratio of encoder tokens in Sparse DETR. Value in the parenthesis indicates the percentage of our high-level tokens compared to the original features. $^\dag$ we adopt the result from the official Deformable DETR codebase. 
    % The meaning of different model variants is described in Sec. \ref{setup}.
    % The models with superscript $^{*}$ use $3$ pattern embeddings as in Anchor DETR~\cite{wang2021anchor}.
    }
    \label{tab:sparsecompare}
    % \vspace{-.3cm}
\end{table*}